\documentclass[10pt,journal,compsoc]{IEEEtran}
\usepackage[nocompress]{cite}
\usepackage[pdftex]{graphicx}
\usepackage{balance}
\graphicspath{{figures/}}
\DeclareGraphicsExtensions{.pdf,.jpeg,.png}
\usepackage{amsmath,bm}
\usepackage{algorithmic}
\usepackage{array}
\usepackage{url}
\usepackage[x11names]{xcolor}
\usepackage{tabulary,xfrac,enumitem}
\usepackage{booktabs}       
\usepackage{amsfonts}       
\usepackage{nicefrac}       
\usepackage{microtype}      
\usepackage{amssymb}
\usepackage{amsmath}
\usepackage{caption, subcaption}
\usepackage{multirow}
\usepackage{textcomp}
\usepackage{makecell}
\usepackage{colortbl}
\usepackage{pifont}
\usepackage{xspace}
\usepackage{subcaption}
\usepackage[ruled,linesnumbered]{algorithm2e}
\usepackage[pagebackref=true,breaklinks=true,colorlinks,citecolor=citecolor,bookmarks=false]{hyperref}
\definecolor{citecolor}{RGB}{34,139,34}

\begin{document}

\title{Bridging the Source-to-target Gap for Cross-domain Person Re-Identification with Intermediate Domains}

\author{Yongxing~Dai,
        Yifan~Sun,
        Jun~Liu,
        Zekun~Tong,
        Yi~Yang,
        Ling-Yu~Duan
\IEEEcompsocitemizethanks{
\IEEEcompsocthanksitem Y. Dai is with the National Engineering Laboratory
for Video Technology, Peking University, Beijing, China. Email: yongxingdai@pku.edu.cn.
Part of this work was done during Yongxing Dai's internship at Baidu Research.
\IEEEcompsocthanksitem Y. Sun is with Baidu Research, Beijing, China. Email: sunyf15@tsinghua.org.cn.
\IEEEcompsocthanksitem J. Liu is with the Information Systems Technology and Design Pillar,
Singapore University of Technology and Design, Singapore. Email: jun\_liu@sutd.edu.sg.
\IEEEcompsocthanksitem Z. Tong is with National University of Singapore, Singapore. Email: zekuntong@u.nus.edu.
\IEEEcompsocthanksitem Y. Yang is with the College of Computer Science and Technology, Zhejiang University, Hangzhou, China. Email: yangyics@zju.edu.cn.
\IEEEcompsocthanksitem L. Duan is with the National Engineering Laboratory for Video
Technology, Peking University, Beijing, China, and also with Peng
Cheng Laboratory, Shenzhen, China. Email: lingyu@pku.edu.cn.
}
\thanks{Corresponding author: Ling-Yu Duan}}

\markboth{WORK IN PROGRESS}%
{}

\IEEEtitleabstractindextext{%
\begin{abstract}

Cross-domain person re-identification (re-ID), such as unsupervised domain adaptive re-ID (UDA re-ID), aims to transfer the identity-discriminative knowledge from the source to the target domain. Existing methods commonly consider the source and target domains are isolated from each other, \emph{i.e.}, no intermediate status is modeled between the source and target domains. Directly transferring the knowledge between two isolated domains can be very difficult, especially when the domain gap is large. This paper, from a novel perspective, assumes these two domains are not completely isolated, but can be connected through a series of intermediate domains. Instead of directly aligning the source and target domains against each other, we propose to align the source and target domains against their intermediate domains so as to facilitate a smooth knowledge transfer. To discover and utilize these intermediate domains, this paper proposes an Intermediate Domain Module (IDM) and a Mirrors Generation Module (MGM). IDM has two functions: 1) it generates multiple intermediate domains by mixing the hidden-layer features from source and target domains and 2) it dynamically reduces the domain gap between the source / target domain features and the intermediate domain features. While IDM achieves good domain alignment effect, it introduces a side effect, \emph{i.e.}, the mix-up operation may mix the identities into a new identity and lose the original identities. Accordingly, MGM is introduced to compensate the loss of the original identity by mapping the features into the IDM-generated intermediate domains without changing their original identity. It allows to focus on minimizing domain variations to further promote the alignment between the source / target domain and intermediate domains, which reinforces IDM into IDM++. We extensively evaluate our method under both the UDA and domain generalization (DG) scenarios and observe that IDM++ yields consistent (and usually significant) performance improvement for cross-domain re-ID, achieving new state of the art. For example, on the challenging MSMT17 benchmark, IDM++ surpasses the prior state of the art by a large margin (\textit{e.g.,} up to 9.9\% and 7.8\% rank-1 accuracy) for UDA and DG scenarios, respectively. Code will be publicly available.

\end{abstract}

}

\maketitle

\IEEEdisplaynontitleabstractindextext

\IEEEpeerreviewmaketitle

\IEEEraisesectionheading{\section{Introduction}\label{sec:introduction}}
\IEEEPARstart{P}{erson} re-identification (re-ID) \cite{zheng2016person,leng2019survey,ye2021deep} aims to identify the same person across non-overlapped cameras.
A critical challenge in a realistic re-ID system is the cross-domain problem, \emph{i.e.}, the training data and testing data are from different domains. Since annotating the person identities is notoriously expensive, it is of great value to transfer the identity-discriminative knowledge from the source to the target domain without incurring additional annotations. To improve the cross-domain re-ID, there are two popular approaches, \emph{i.e.}, the unsupervised domain adaption (UDA) and domain generalization (DG). These two approaches are closely related yet have an important difference: during training, UDA has unlabeled target-domain data, while DG is not accessible to the target domain and is thus more challenging. This paper mainly challenges the UDA re-ID and provides compatibility to DG re-ID.  

Existing UDA re-ID methods commonly consider the source and target domains are isolated from each other, \emph{i.e.}, there is no intermediate status between the source and target domains. Directly transferring knowledge between two isolated domains can be very difficult, especially when the domain gap is large.
Specifically, there are two approaches, \emph{i.e.}, style transfer~\cite{wei2018person,deng2018image} and pseudo-label-based training~\cite{song2020unsupervised,fu2019self,dai2020dual,ge2020self}. 
The style-transfer methods usually use GANs \cite{zhu2017unpaired} to transfer the target-domain style onto the labeled source-domain images, so that the deep model can learn from the target-stylized images. Since the source and target domains are isolated and far away from each other, the style transfer procedure can be viewed as jumping from the source to the target domain, which can be difficult. 
The pseudo-label-based training methods require a clustering procedure as the prerequisite for obtaining the pseudo labels. The domain gap between the isolated source and target domains  compromises the clustering accuracy and incurs noisy pseudo labels.
Therefore, we conjecture that directly mitigating the domain gap between two isolated (source and target) domains is difficult for both the style-transfer and the pseudo-label-based UDA approaches.

\begin{figure}[tp]
\begin{center}
\includegraphics[width=1.0\linewidth]{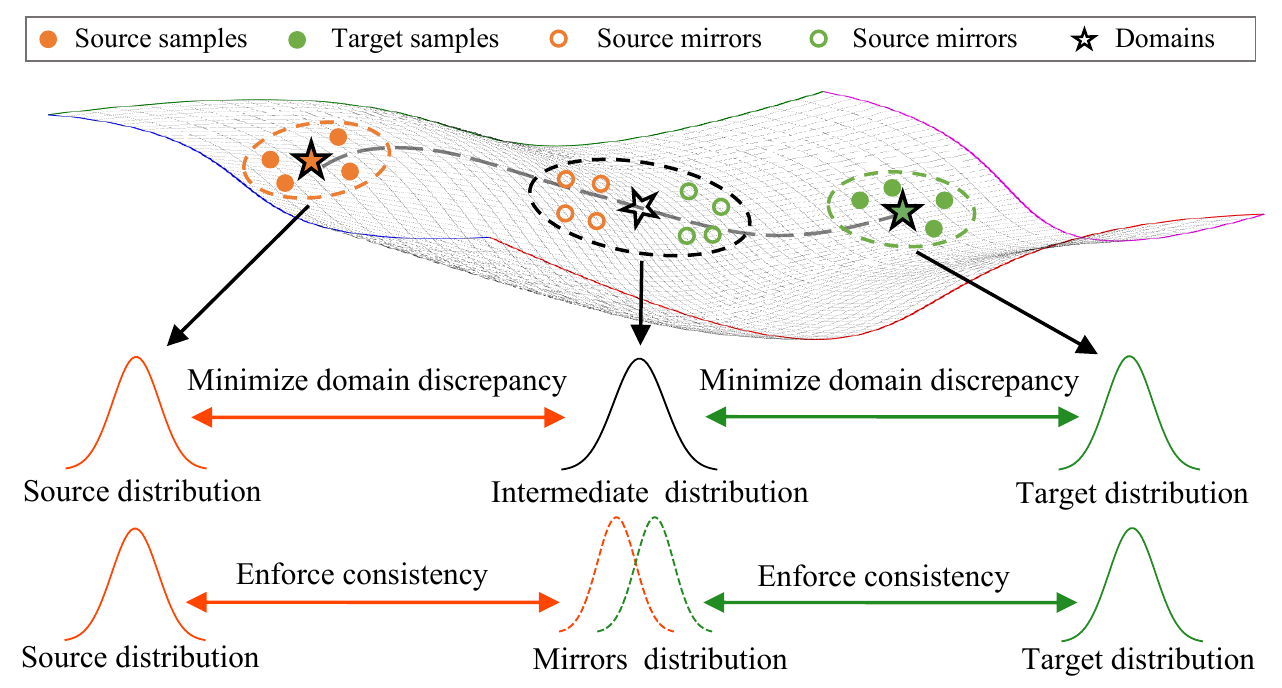}
\end{center}
\caption{Illustration of our main idea. Assuming that the source and target domains (in UDA re-ID) are located in a manifold, there can be some intermediate domains along with the path to bridge the two extreme domains. 
With the generation of intermediate domains, the source and target domains can be smoothly aligned with them.
To further preserve the source / target identity information during alignment, we map source / target identities into the intermediate domains to obtain source / target mirrors.
}
\label{fig:intro}
\end{figure}

From a novel perspective, this paper considers that the source and target domains are not isolated, but are potentially connected through a series of intermediate domains. 
In other words, some intermediate domains underpin a path that bridges the gap between the source and target domains. 
Specifically, we assume that the source and target domains are located in a manifold, as shown in Fig.~\ref{fig:intro}. There is an appropriate ``path'' connecting these two isolated domains.  
The source and target domains lie on the two extreme points of this path, while some intermediate domains exist along this path, characterizing the inter-domain connections. This viewpoint leads to an intuition: to align the two extreme points, comparing them against these intermediate points can be more feasible than directly comparing the two extreme points against each other, given that they are far from each other due to the significant domain gap.

Motivated by this, we propose an intermediate domain module (IDM) for UDA re-ID. Instead of directly aligning the source and target domain against each other (\emph{i.e.}, source $\rightarrow$ target), IDM aligns the source and target domain against a shared set of intermediate domains (\emph{i.e.}, source $\rightarrow$ intermediate and target $\rightarrow$ intermediate), as illustrated in Fig.~\ref{fig:intro}. To this end, IDM first synthesizes multiple intermediate domains and then minimizes the distance between each intermediate domain and the source / target domain. Specifically, IDM mixes the source-domain and target-domain features, which are output from a hidden layer of the deep network. The correspondingly-synthesized features exhibit the characteristics of the intermediate domain and are then fed into the sub-sequential hidden layer. The consequential output features of the intermediate domains are compared against the original source and target domain features for minimizing the domain discrepancy. We note that the IDM can be plugged after any hidden layers and is thus a plug-and-play module. There are two critical techniques for constructing the IDM:

$\bullet$ \emph{Modeling the intermediate domain.}
We use the Mixup \cite{zhang2017mixup} strategy to mix the source and target hidden features (\emph{e.g.}, ``Stage-0'' in ResNet-50) and forward the mixed hidden feature until the last deep embedding. Importantly, we consider the mix ratio in the hidden layer determines the position of the output feature in the deeply-learned embedding space. More concretely, if the mixup operation in the hidden feature space assigns a larger proportion to the source domain, the output feature should be closer to the source domain than to the target domain (and vice versa). Correspondingly, when we minimize its distance to the source and target domain in parallel, we will emphasize both domains proportionally. In a word, the correspondence between the mix ratio and the relative distance on the path connecting the source and target domain models the intermediate domain. 

$\bullet$ \emph{Enhancing the intermediate domain diversity.}
The path between the source and target domains is jointly depicted by all the intermediate domains on this path. Therefore, it is important to generate the intermediate domains densely. Since the mix ratio determines the position of the mixed features on the path, varying the mix ratio results in different intermediate domains. IDM adaptively generates the mix ratios and encourages the mix ratios to be as diverse as possible for different input hidden features.

Based on the IDM in our previous conference version~\cite{dai2021idm}, this paper makes an important extension, \emph{i.e.}, reinforcing IDM into IDM++ by further promoting the alignment from source / target domain to intermediate domains. In IDM, when a source and a target-domain feature are mixed up, their identities are accordingly mixed into a new identity (Section~\ref{sec:IDM-modelling}). This mixup has a side effect, \emph{i.e.}, there are no original identities in the intermediate domains. Therefore, the cross-domain alignment by IDM may focus more on the mixed-identity variation than the diverse cross-domain styles that are irrelevant to the identity.
To compensate this, we propose a Mirrors Generation Module (MGM) to map original identities into the already-generated intermediate domains. MGM utilizes a popular statistic-based style transfer approach AdaIN~\cite{huang2017arbitrary}, which replaces the mean and standard deviation value of the source / target features with those from intermediate features. Consequently, each feature in the source / target domain has a ``same-identity different-style'' mirror in every intermediate domain, as shown in Fig.~\ref{fig:intro}. 
It enables IDM++ to focus on the cross-domain variation when aligning the domains (as detailed in Section~\ref{sec:domain_mapping}).
Thus, IDM++ brings another round of significant improvement based on IDM.

We note that although the statistic-based style transfer is not new and is widely adopted by many recent out-of-distribution literature \cite{zhou2021domain,nuriel2021permuted,tang2021crossnorm}, integrating MGM implemented with style transfer into IDM++ brings novelty and significant benefits. First, the style transfer in prior methods is between off-the-shelf domains, while our MGM is the first to conduct style transfer towards on-the-fly synthesized domains (\emph{i.e.}, the IDM-generated intermediate domains). 
Second, on UDA re-ID, we show that MGM and IDM mutually reinforce each other and jointly bring complementary benefits for IDM++. On the one hand, adding MGM significantly improves IDM (\emph{e.g.}, 8.6\% improvement of rank-1 accuracy on Market-1501$\to$MSMT17). On the other hand, IDM is a critical prerequisite for the effectiveness of MGM, because we find that using MGM without IDM only brings slight improvement. Third, MGM endows IDM++ with the extra capacity of improving domain generalization, allowing IDM++ to be the first framework to integrate both UDA and DG capacity for cross-domain re-ID, achieving state of the art results on both scenarios. 

In addition to the reinforcement from IDM to IDM++, this paper makes two more extensions \emph{w.r.t.} the applied scenarios and the experimental evaluation. Overall, compared with our previous conference version~\cite{dai2021idm}, the extensions are from three aspects: \emph{1) Method.} We reinforce IDM into IDM++ by adding a novel MGM (Section~\ref{sec:domain_mapping}). MGM maps the training identities from the source / target domain as their mirrors in the IDM-generated intermediate domains. It allows IDM++ to directly minimize the variation of the same identity across various domains. This new advantage improves IDM by a large margin.
\emph{2) Applied scenarios.}
We investigate IDM++ under two popular cross-domain re-ID scenarios, \emph{i.e.}, unsupervised domain adaptation (UDA) and the more challenging domain generalization (DG). We show that IDM++ provides a unified framework for these two scenarios and brings general improvement. \emph{3) Experimental evaluation.} We provide more comprehensive experimental evaluation, including experiments on more detailed ablation studies (Section~\ref{sec:ablation}), analysis on parameter sensitivity (Section~\ref{sec:parameter}), two challenging protocols of DG re-ID (Section~\ref{sec:SOTA}), investigations on key designs of IDM++ (Section~\ref{sec:investigation_on_mechanism}), and discussions on the distribution along the domain bridge and alignment of domains (Section~\ref{sec:analysis}).

Our contributions can be summarized as follows.
\begin{itemize}
    \item To the best of our knowledge, we are the first to explore intermediate domains for the cross-domain re-ID task. 
    We consider that some intermediate domains can be densely distributed between the source / target domain and jointly depict a smooth bridge across the domain gap. Minimizing the domain discrepancy along this bridge improves domain alignment.

     \item We propose an IDM module to discover the desired intermediate domains by on-the-fly mixing the source and target domain features with varying mix-ratios. Afterwards, IDM mitigates the gap from source / target to the intermediate domains so as to align the source and target domains.

    \item We propose an MGM module to further promote the source / target to intermediate domain alignment.
    In compensation for a loss of IDM (\emph{i.e.}, the feature mix-up loses the original training identities), MGM maps the original training identities into the intermediate domains. Incorporating MGM and IDM, IDM++ brings another round of substantial improvement.
    
    \item We show that our IDM++ provides a unified framework for two popular cross-domain re-ID scenarios, \emph{i.e.}, UDA and DG. 
    Extensive experiments under twelve UDA re-ID benchmarks and two DG re-ID protocols validate that IDM++ brings general improvement and sets new state of the art for both scenarios.
    
\end{itemize}

\section{Related Work}\label{sec:related_work}
\textbf{Unsupervised Domain Adaptation.} 
A line of works converts the UDA into an adversarial learning task \cite{ganin2015unsupervised,tzeng2017adversarial,long2018conditional,russo2018source}. 
Another line of works uses various metrics to measure and minimize the domain discrepancy, such as MMD \cite{long2015learning} or other metrics \cite{sun2016deep,zhuo2017deep,long2017deep,kang2018deep}.
Another line of traditional works \cite{gong2012geodesic,cui2014flowing,gopalan2013unsupervised} tries to bridge the source and target domains based on intermediate domains. In traditional methods \cite{gong2012geodesic,gopalan2013unsupervised}, they embed the source and target data into a Grassmann manifold and learn a specific geodesic path between the two domains to bridge the source and target domains. Still, they are not easily applied to the deep models. In deep methods \cite{gong2019dlow,cui2020gradually}, they either use GANs to generate a domain flow by reconstructing input images on pixel level\cite{gong2019dlow} or learn better domain-invariant features by bridging the learning of the generator and discriminator \cite{cui2020gradually}. 
However, reconstructing images may not guarantee the high-level domain characteristics in the feature space or introduce unnecessary noise in the pixel space. It will become harder to adapt between the two isolated domains especially when the large domain gap.
Unlike the above methods, we propose a lightweight module to model the intermediate domains, which can be easily inserted into the existing deep networks. Instead of hard training for GANs or reconstructing images, our IDM can be learned in an efficient joint training scheme. Besides, this paper also proposes a new module (MGM) to encourage the model to focus on minimizing the cross-domain variation when aligning the source and target domains.

\textbf{Unsupervised Domain Adaptive Person Re-ID.} In recent years, many UDA re-ID methods have been proposed and they can be mainly categorized into three types based on their training schemes, \textit{i.e.,} GAN transferring \cite{wei2018person,deng2018image,huang2019sbsgan,zou2020joint}, fine-tuning \cite{song2020unsupervised,fu2019self,ge2020mutual,dai2020dual,chen2020deep,zhai2020multiple,jin2020global,lin2020unsupervised,zhai2020ad,wang2022attentive}, and joint training \cite{zhong2019invariance,zhong2020learning,wang2020unsupervised,ge2020self,ding2020adaptive,zheng2021online,isobe2021towards}. 
GAN transferring methods use GANs to transfer images' style across domains \cite{wei2018person,deng2018image} or disentangle features into id-related/unrelated features \cite{zou2020joint}.
For fine-tuning methods, they first train the model with labeled source data and then fine-tune the pre-trained model on target data with pseudo labels. 
The key component of these methods is how to alleviate the effects of the noisy pseudo labels. 
However, these methods ignore the labeled source data while fine-tuning on the target data, which will hinder the domain adaptation process because of the catastrophic forgetting in networks. For joint training methods, they combine the source and target data together and train on an ImageNet-pretrained network from scratch. 
All these joint training methods often utilize the memory bank \cite{xiao2017joint,wu2018unsupervised} to improve target domain features' discriminability. However, these methods just take both the source and target data as the network's input and train jointly while neglecting the bridge between both domains, \textit{i.e.,} what information of the two domains' dissimilarities/similarities can be utilized to improve features' discriminability in UDA re-ID.
Different from all the above UDA re-ID methods, we propose to consider the bridge between the source and target domains by modeling appropriate intermediate domains with a plug-and-paly module, which is helpful for gradually adapting between two extreme domains in UDA re-ID.
Besides, we also propose a novel statistic-based feature augmentation module to exploit the intermediate domains' styles, which can encourage the model to focus on learning identity-discriminative features when aligning the domains. The above exploitation of intermediate domains has not been investigated in this field as far as we know.

\textbf{Mixup and Variants.} Mixup \cite{zhang2017mixup} is an effective regularization technique to improve the generalization of deep networks by linearly interpolating the image and label pairs, where the interpolating weights are randomly sampled from a Dirichlet distribution. Manifold Mixup \cite{verma2019manifold} extends Mixup to a more general form which can linearly interpolate data at the feature level. Recently, Mixup has been applied to many tasks like point cloud classification \cite{chen2020pointmixup}, object detection \cite{zhang2019bag}, and closed-set domain adaptation\cite{xu2020adversarial,wu2020dual,na2021fixbi}. Our work differs from these Mixup variants in: (1) All the above methods take Mixup as a data/feature augmentation technique to improve models' generalization, while we bridge two extreme domains by generating intermediate domains for cross-domain re-ID. (2) We design an IDM module and enforce specific losses on it to control the bridging process while all the above methods often linearly interpolate data using the random interpolation ratio without constraints.

\textbf{Domain Generalization.} Domain generalization (DG) aims to improve the model's generalization on one or more target domains by only using one or more source domains for training. Different from UDA, target data is not accessible for training DG models. For more comprehensive surveys of DG, please refer to \cite{zhou2021domain2,wang2021generalizing,shen2021towards}. Following the taxonomy in a DG survey \cite{zhou2021domain2}, we briefly review the two categories that are most related to our work: 1) domain alignment and 2) data augmentation. The key idea of the existing domain alignment methods is to align different source domains by only minimizing their discrepancy.
Different from these works, we align the source and target domains by aligning them with the synthetic intermediate domain respectively. The motivation of our work is to ease the procedure of domain alignment using intermediate domains, which fundamentally differs from the existing works. Data augmentation works in DG aim to improve models' generalization by designing image-based \cite{volpi2018generalizing,qiao2020learning,zhou2020learning,zhou2020deep} or feature-based \cite{mancini2020towards,xu2021robust,zhou2021domain,nuriel2021permuted,tang2021crossnorm} augmentation methods, which can avoid over-fitting to source domains. For example, several feature-based augmentation researches \cite{zhou2021domain,nuriel2021permuted,tang2021crossnorm} focus on manipulating feature statistics based on the observation that CNN feature statistics (\textit{i.e.,} mean and standard deviation) can represent an image's style \cite{huang2017arbitrary}. Motivated by them, we propose a new augmentation module (\textit{i.e.,} MGM) implemented by the arbitrary-style-transfer technique AdaIN \cite{huang2017arbitrary} to augment source/target features with statistics of the already-generated intermediate domains. Different from the existing augmentation works, in this paper, we exploit the diverse inter-domain variation to augment source / target identities with different domain styles, and we utilize this augmentation mechanism to complement with the procedure of generating intermediate domains by our IDM.

\textbf{Domain Generalizable Person Re-ID.} Recently, a newly proposed cross-domain re-ID task called domain generalizable re-ID (DG re-ID) \cite{song2019generalizable,jia2019frustratingly,jin2020style,zhao2021learning,choi2021meta,dai2021generalizable} has gained much interest of researchers. DG re-ID aims to improve the generalization on the target domains that are not accessible during training, where only one or more source domains can be used for training. Different from the classification task in DG, the target domain does not share the label space with the source domains, which poses a new challenge over conventional DG settings. The existing DG re-ID works mainly fall into two categories: 1) designing specific modules that can learn better domain-invariant representations \cite{song2019generalizable,jia2019frustratingly,jin2020style}, and 2) adopting the meta-learning paradigm to simulate the domain bias by splitting training domains into pseudo-seen and pseudo-unseen domains \cite{zhao2021learning,choi2021meta,dai2021generalizable}. For example, SNR \cite{jin2020style} designs a Style Normalization and Restitution module to disentangle object representations into identity-irrelevant and identity-relevant ones. RaMoE \cite{dai2021generalizable} introduces meta-learning into a novel mixture-of-experts paradigm via an effective voting-based mixture mechanism, which can learn to dynamically aggregate multi-source domains information. Different from the above works, we aim to exploit the bridge among different source domains to tackle the problem of DG re-ID, which may provide a novel perspective in solving this problem.

\section{Methodology}\label{sec:methodology1}
This paper explores intermediate domains for cross-domain re-ID. We first focus on the unsupervised domain adaptation (UDA) scenario (from Section \ref{sec: overview} to \ref{sec:domain_mapping}) and then extend our method to the domain generalization (DG) scenario (Section \ref{sec:extension_dg_reid}). Specifically, for UDA, we first introduce the overall pipeline of IDM++ in Section \ref{sec: overview}, and then elaborate the key component IDM for discovering the intermediate domains in Section \ref{sec:IDM-module}. 
Given the learned intermediate domains, we disentangle the source-to-target domain alignment into a more feasible one, \emph{i.e.}, aligning source / target to intermediate domains in Section \ref{sec:bridge_loss}. Afterwards, we further improve the source / target to intermediate domain alignment with MGM and summarize the overall training for IDM++ in Section \ref{sec:domain_mapping}. Section \ref{sec:extension_dg_reid} illustrates how to apply IDM++ for DG re-ID.

\begin{figure*}[htp]
\begin{center}
\includegraphics[width=1.0\linewidth]{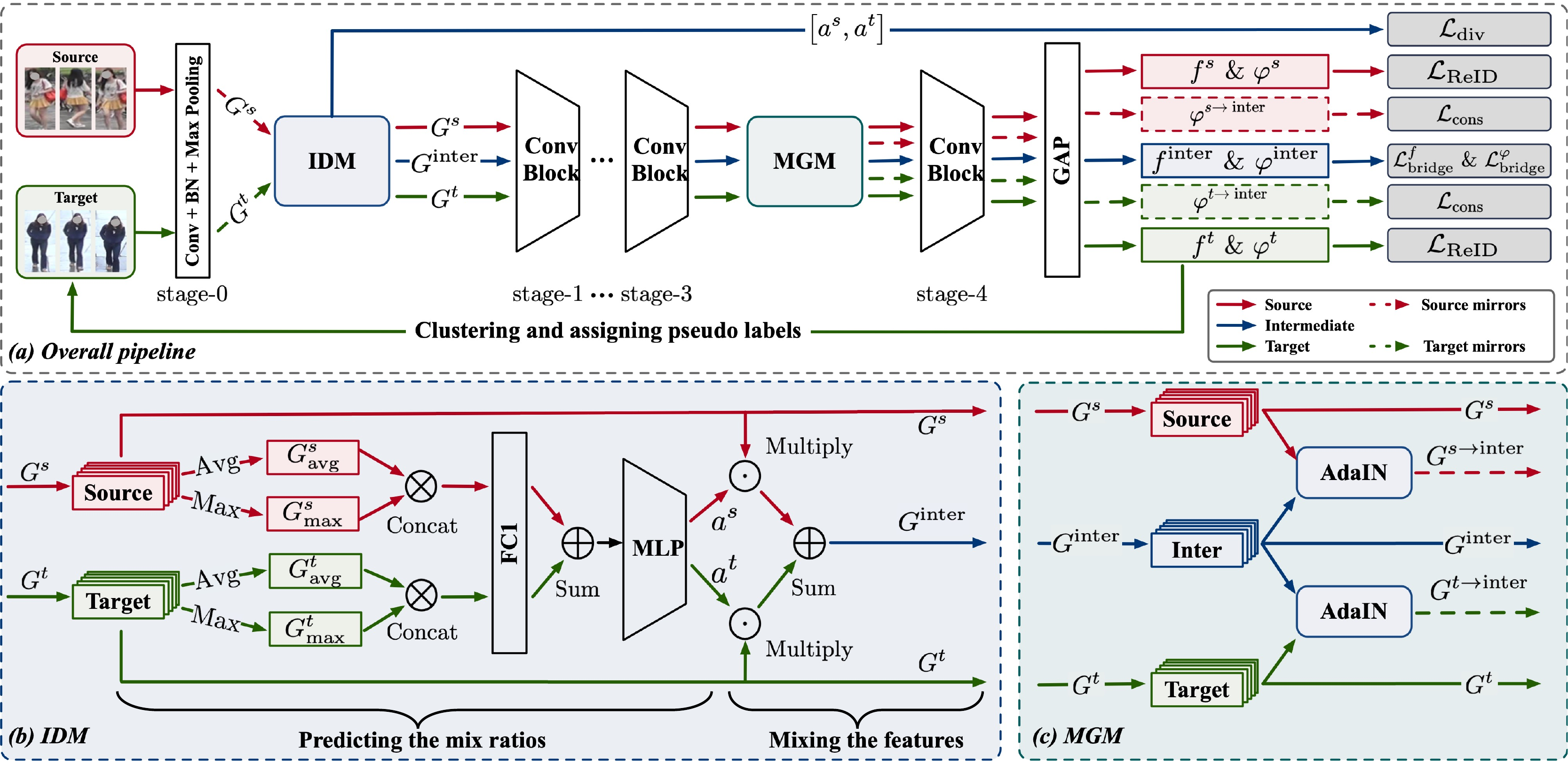}
\end{center}
\caption{The overall of IDM++ (a), which is comprised of IDM (b) and MGM (c). In (a), IDM++ combines the source and target domain for joint training. It first plugs IDM after a bottom layer (\emph{e.g.}, stage-0 in ResNet-50~\cite{he2016deep})  and then plugs MGM after a top layer (\emph{e.g.}, stage-3). IDM mixes the hidden-layer features from the source and target domain (\textit{i.e.,} $G^{s}, G^{t}$) to generate intermediate-domain features $G^{\rm inter}$. $G^{s}$, $G^{t}$, and $G^{\rm inter}$ continue to proceed along the network until the final outputs, \emph{i.e.}, the deep embedding ($f^{s}$, $f^{t}$ and $f^{\rm inter}$) and the softmax prediction ($\varphi^{s}$, $\varphi^{t}$ and $\varphi^{\rm inter}$). Since the intermediate-domain features are initially generated after stage-0, we consider the following stages already provide intermediate domains. Given the intermediate-domains features at the end of stage-3, MGM maps the features in the source and target domain into the intermediate domains, \emph{i.e.}, $G^{s \to \rm inter}$ and $G^{t \to \rm inter}$, which are finally output as $\varphi^{s \to \rm inter}$  and $\varphi^{t \to \rm inter}$ in the softmax prediction layer. Overall, there are four losses for learning IDM++: 1) Given the output of the source and target domain, we enforce the popular ReID loss $\mathcal L_{\rm ReID}$ (including the classification loss $\mathcal L_{\rm cls}$ and triplet loss $\mathcal L_{\rm tri}$). 2) Given the output of IDM, \emph{i.e.}, $f^{\rm inter}$ and $\varphi^{\rm inter}$ in the intermediate domain, we enforce two bridge losses \textit{i.e.,} $\mathcal{L}^{f}_{\rm bridge}$ and $\mathcal{L}^{\varphi}_{\rm bridge}$. 3) Given the output of MGM, \emph{i.e.}, $\varphi^{s \to \rm inter}$ and $\varphi^{t \to \rm inter}$ in the intermediate domain, we enforce the consistency loss $\mathcal{L}_{\rm cons}$. 4) To generate more intermediate domains, we use $\mathcal{L}_{\rm div}$ to enlarge the variance of the mix ratio.}
\label{fig:pipeline}
\vspace{-1em}
\end{figure*}

\subsection{Overview}\label{sec: overview}
In UDA re-ID, we are often given a labeled source domain dataset $\{(x^{s}_{i},y^{s}_{i})\}$ and an unlabeled target domain dataset $\{x^{t}_{i}\}$. The source dataset contains $N^{s}$ labeled person images, and the target dataset contains $N^{t}$ unlabeled images. Each source image $x_{i}^{s}$ is associated with a person identity $y_{i}^{s}$ and the total number of source domain identities is $C^{s}$. We adopt the pseudo-label-based pipeline to perform clustering to assign the pseudo labels for the target samples. Similar to the pseudo-label-based UDA re-ID methods~ \cite{fu2019self,song2020unsupervised,ge2020self}, we perform DBSCAN clustering on the target domain features at the beginning of every training epoch and assign the cluster id (a total of $C^{t}$ clusters) as the pseudo identity $y^{t}_{i}$ of the target sample $x^{t}_{i}$. We use ResNet-50 \cite{he2016deep} as the backbone network $f(\cdot)$ and add a hybrid classifier $\varphi(\cdot)$ after the global average pooling (GAP) layer, where the hybrid classifier is comprised of the batch normalization layer and a $C^{s}+C^{t}$ dimensional fully connected (FC) layer followed by a softmax activation function.

Fig.~\ref{fig:pipeline}~(a) shows the overall framework of our method. 
Our IDM++ adopts the source-target joint training pipeline, \emph{i.e.}, each mini-batch consists of $n$ source samples and $n$ target samples. IDM++ has two key components, \emph{i.e.}, IDM and MGM. Both IDM and MGM can be seamlessly plugged into the backbone network (\emph{e.g.}, Resnet-50), while IDM should be placed on earlier layers because it is a prerequisite for MGM. 
An optimized configuration is to plug IDM between the stage-0 and stage-1 and plug MGM between the stage-3 and stage-4. 
The IDM generates intermediate domains by mixing the source-domain and target-domain feature with two corresponding mix ratios, \emph{i.e.}, $a^{s}$ and $a^{t}$. The mixed features, along with the original source-domain and target-domain features in the hidden layer continue to proceed until the deep embedding, \emph{i.e.}, the features after the global average pooling (GAP), and correspondingly become $f^{\rm inter}$, $f^{s}$ and $f^{t}$. With the classifier, these features are mapped to $\varphi^{\rm inter}$, $\varphi^{s}$ and $\varphi^{t}$ respectively.
The MGM maps the original source / target identities into the IDM-generated intermediate domains to obtain the mirrors. After the classifier, we obtain the source and target mirrors' predictions: $\varphi^{s \to \rm inter}$ and $\varphi^{t \to \rm inter}$.

\subsection{Intermediate Domain Module}
\label{sec:IDM-module}
In this section, we first illustrate how to utilize IDM to generate intermediate domains' features in the hidden stage of the backbone (Section~\ref{sec:IDM-modelling}). Next, we provide the manifold assumption to exploit two properties (proportional distance relationship in Section~\ref{sec:IDM-proportional} and diversity in Section~\ref{sec:IDM-diversity}) that the intermediate domains should satisfy in the deeply-learned output space.

\subsubsection{Modelling the intermediate domains}
\label{sec:IDM-modelling}
We denote backbone network as $f(x)=f_{m}(G)$, where $G$ is the hidden feature map $G \in \mathbb{R}^{h\times w\times c}$
after the $m$-th stage and $f_{m}$ represents the part of the network mapping the hidden representation $G$ after the $m$-th hidden stage to the 2048-dim feature after the GAP layer.
As shown in Fig.~\ref{fig:pipeline}~(b), IDM contains two steps: 1) \textbf{predicting the mix ratios} and 2) \textbf{mixing the features}. Specifically, IDM is plugged after a bottom (earlier) stage in the backbone network and takes the hidden-layer features from both the source and target domain ($G^s$ and $G^t$) as its input to predict the two mix ratios (\emph{i.e.}, $a^s$ and $a^t$). Next, IDM mixes $G^s$ and $G^t$ with two corresponding mix ratios to generate intermediate domains hidden-layer features $G^{\rm inter}$. The $G^{s}$, $G^{t}$ and $G^{\rm inter}$ forward into the next hidden stage until the final output of the network.

\textbf{Predicting the mix ratios.} As shown in Fig.~\ref{fig:pipeline}~(b), 
in a mini-batch comprised of $n$ source and $n$ target samples, we randomly assign all the samples into $n$ pairs so that each pair contains a source-domain and a target-domain sample.
For each sample pair $(x^{s},x^{t})$, the network obtains two feature maps at the $m$-th hidden stage: $G^{s}, G^{t} \in \mathbb{R}^{h\times w\times c}$.
IDM uses an average pooling and a max pooling operation to transform each single feature into two $1\times 1 \times c$ dimensional features. Therefore, for each source-target pair, we have ($G_{avg}^{s}, G_{max}^{s}$) for the source domain, and ($G_{avg}^{t}, G_{max}^{t}$) for the target domain.
IDM concatenates the $avg$-pooled and $max$-pooled features for each domain and feed them into a fully-connected layer, \emph{i.e.}, FC1 in Fig.~\ref{fig:pipeline} (b). The output feature vectors of FC1 are merged using element-wise summation and fed into a multi-layer perception (MLP) to obtain a mix ratio vector $a =[a^{s},a^{t}] \in \mathbb{R}^{2}$. The above step for predicting the mix ratios is formulated as:
\begin{equation}
\small
a=\delta(MLP(FC1([G_{avg}^{s};G_{max}^{s}])+FC1([G_{avg}^{t};G_{max}^{t}]))),
\label{eq:attention}
\end{equation}
where $\delta(\cdot)$ is the softmax function to ensure $a^{s}+a^{t}=1$.

\textbf{Mixing the features.} 
Given the predicted mix ratios $a^s$ and $a^t$, IDM mixes the source-domain and target-domain features ($G^s$ and $G^t$) using weighted mean operation, as illustrated in Fig.~\ref{fig:pipeline}~(b). The mixing step is formulated as:
\begin{equation}
\small
    G^{\rm inter}=a^{s}\cdot G^{s}+a^{t}\cdot G^{t},
\label{eq:Mix}
\end{equation}

These hidden features (\textit{i.e.,} $G^{s}$, $G^{t}$, and $G^{\rm inter}$) proceed until the outputs (\textit{e.g.,} embeddings after the GAP layer or logits after the classifier in re-ID) of the network.
We note a side-effect accompany the above mixing step, \emph{i.e.}, the mixing of identities. Concretely, we consider the mixed feature $G^{\rm inter}$ has a new identity, \textit{i.e.}, $y^{\rm inter}=a^{s}\cdot y^{s}+a^{t} \cdot y^{t}$.

\subsubsection{Proportional distance relationship}
\label{sec:IDM-proportional}
\begin{figure}[htp]
\begin{center}
\includegraphics[width=0.9\linewidth]{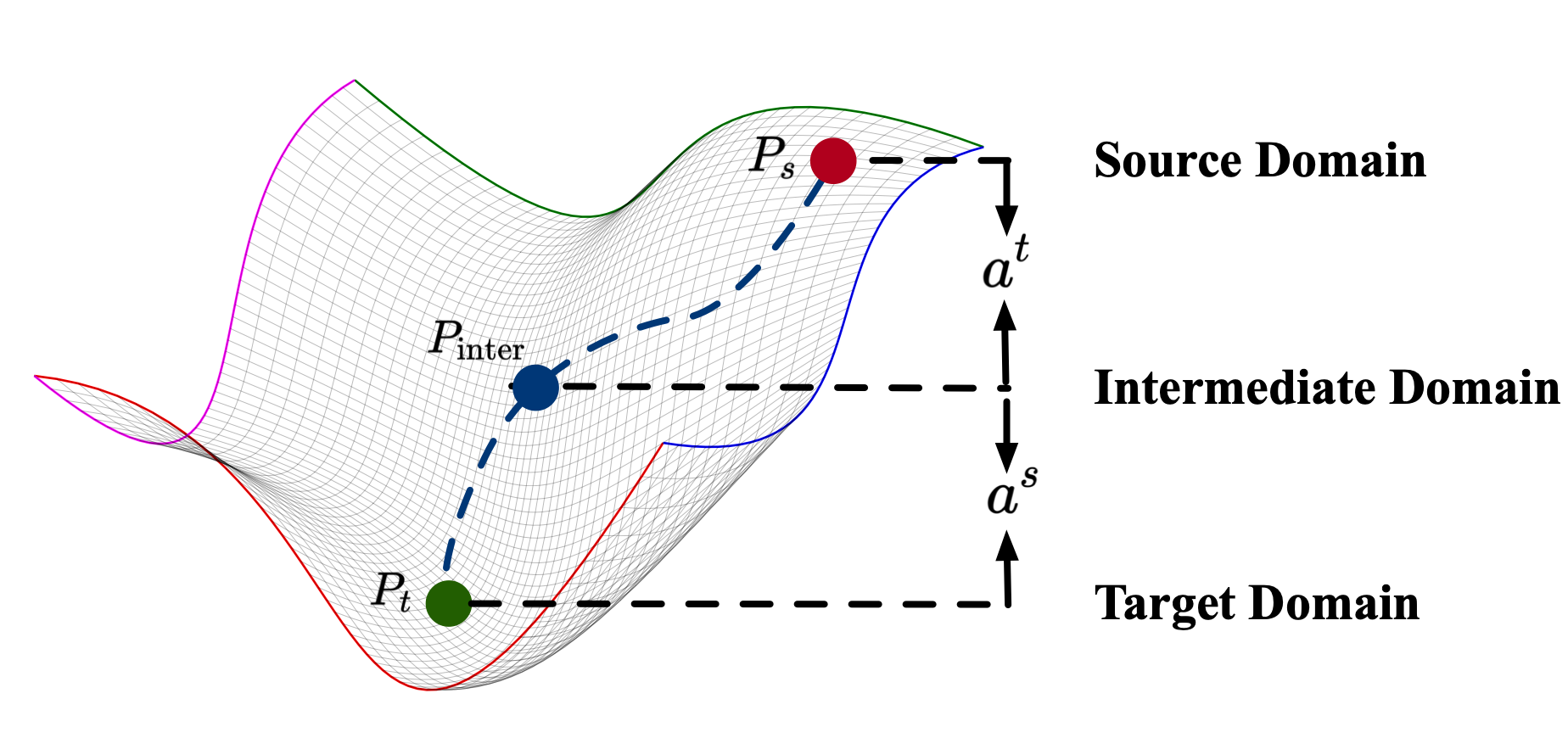}
\end{center}
\caption{Assuming domains as points in a manifold. The ``shortest geodesic path'' connecting the source and target domains (\textit{i.e.,} $P_{s}$ and $P_{t}$) is bridged by dense intermediate domains (\textit{i.e.,} $P_{\rm inter}$).
We use the mix ratios (\textit{i.e.,} $a^{s}$ and $a^{t}$) to approximate the distance relationship between domains in the manifold.}
\label{fig:manifold}
\end{figure}

We assume that the source and target domains are located on a manifold, \emph{i.e.}, a topological space that locally resembles Euclidean space. On a manifold, the distance between two isolated and far-away points cannot be measured through Euclidean distance. Instead , the distance should be measure through the ``shortest geodesic distance'' \cite{gong2012geodesic,gopalan2013unsupervised}, \textit{i.e.,} the curve representing the shortest path bridging two isolated points in a manifold surface (as shown in Fig.~\ref{fig:manifold})

\textbf{Definition (Shortest geodesic distance).}
To estimate such ``shortest geodesic distance'', we assume there exist many intermediate domains that are close to both the source and target domains. In other words, we assume that the distance between the intermediate domain and the source / target domain is small enough that they may be viewed as lying in a local Euclidean space. Therefore, we sum up the source-to-intermediate Euclidean distance and the target-to-intermediate Euclidean distance as the shortest geodesic distance between the source and target domain samples, which is formulated as:
\begin{equation}
d(P_{s},P_{t}) = d(P_{s},P_{\rm inter})+d(P_{t},P_{\rm inter}).
\end{equation}
where the $P_s$, $P_t$ and $P_{\rm inter}$ are the source domain, the target domain and the intermediate domain, respectively. In IDM, when the mixed feature $G^{\rm inter}$ (Eq.~(\ref{eq:Mix})) proceeds along the sub-sequential layers, we view all the corresponding features in the sub-sequential layers are within the intermediate domain. Based on the above definition, the objective of aligning the source against target domain can be transformed into simultaneously aligning the source and target domain against the intermediate domains.

In Eq.~(\ref{eq:Mix}), the mix ratios $a^s$ and $a^t$ jointly control the relevance of the intermediate domains to the source and target. If the source ratio $a^{s}$ is larger than the target ratio $a^{t}$, the synthetic intermediate domain should be closer to the source domain than to the target domain. Therefore, we utilize the mix ratio to approximately estimate the distance relationship between the intermediate domains' outputs and other two isolated domains' outputs. Specifically, we formulate such distance relationship of these domains' outputs as the following \textbf{Property-1}.

\textbf{Property-1 (Distance should be proportional):}
\textit{We use the mix ratios (\textit{i.e.,} $a^{s}$ and $a^{t}$) to approximate the distance relationship between the intermediate domains' output $P_{\rm inter}$  and the other two isolated domains' output (\textit{i.e.,} $P_{s}$ and $P_{t}$). The distance relationship is formulated as follows:}
\begin{equation}
\small
\frac{d(P_{s},P_{\rm inter})}{d(P_{t},P_{\rm inter})}=\frac{a^{t}}{a^{s}}=\frac{1-a^{s}}{a^{s}}=\frac{a^{t}}{1-a^{t}}.
\label{eq:proportional2}
\end{equation}

Using the proportional distance relationship in Eq.~(\ref{eq:proportional2}), we can approximately locate the position of the intermediate domains in the deeply-learned embedding space or prediction space.

\subsubsection{Promoting the diversity of intermediate domains}
\label{sec:IDM-diversity}

As shown in the ``shortest geodesic distance'' definition, the two isolated domains (\textit{i.e.,} source and target) cannot be directly measured in the Euclidean space if they are far away from each other. The distance of two distant points must be measured by the points existing between them. To estimate the ``shortest geodesic distance'' more precisely, we need as many as possible intermediate points. Compared with sparse intermediate points, more dense points can characterize the "shortest geodesic path" bridging the source and target domains more comprehensively.  
Therefore, we propose the \textbf{Property-2} as follows.

\textbf{Property-2 (Diversity):}
\textit{Intermediate domains should be as diverse as possible.}

Based on the above property, the distance of the source and target domains (\textit{i.e.,} $d(P_{s},P_{t})$) can be correctly measured in the manifold by approximating the distance between the dense intermediate domains and the other two isolated domains (\textit{i.e.,} $d(P_{s},P_{\rm inter})$ and $d(P_{t},P_{\rm inter})$). To satisfy such property when generating intermediate domains in training, we propose a diversity loss by maximizing the differences of the mix ratios ($a^{s}$ and $a^{t}$) within a mini-batch. This loss is formulated as follows:
\begin{equation}
\small
\mathcal L_{\rm div}=-[\sigma(\{a^{s}_{i}\}_{i=1}^{n})+\sigma(\{a^{t}_{i}\}_{i=1}^{n})],
\label{eq:loss_div}
\end{equation}
where $\sigma(\cdot)$ means calculating the standard deviation of the values in a mini-batch. By minimizing $\mathcal L_{\rm div}$, we can enforce intermediate domains to be as diverse as enough to model the characteristics of the ``shortest geodesic path'', which can better bridge the source and target domains.

\subsection{Aligning by Intermediate Domains}
\label{sec:bridge_loss}
Instead of directly aligning the source and target domains (\textit{i.e.,} source $\to$ target), we utilize the IDM to align the source and target domains against the synthetic intermediate domains (\textit{i.e.,} source $\to$ intermediate and target $\to$ intermediate), as illustrated in Fig.~\ref{fig:intro}. Given the ``shortest geodesic distance'' definition and \textbf{Property-1} in Section~\ref{sec:IDM-modelling}, we propose the bridge loss to adaptively minimize the discrepancy between the IDM-generated intermediate domains' output (\textit{i.e.,} $P_{\rm inter}$) and the source and target domains' output (\textit{i.e.,} $P_{s}$ and $P_{t}$). The general form of the bridge loss is formulated as follows:
\begin{equation}
\small
\mathcal L_{\rm bridge}=a^{s}\cdot d(P_{s},P_{\rm inter})+a^{t}\cdot d(P_{t},P_{\rm inter}).
\label{eq:loss_mix}
\end{equation}

The intuition of the $\mathcal{L}_{\rm bridge}$ is based on \textbf{Property-1}. As shown in \textbf{Property-1}, we utilize the proportional relationship (\textit{i.e.,} Eq.~(\ref{eq:proportional2})) to estimate the location of the IDM-generated intermediate domains along the domain bridge. If the intermediate domain $P_{\rm inter}$ is closer to the source $P_{s}$ than the target $P_{t}$ (\textit{i.e.,} $a^{s}$ is larger than $a^{t}$), the objective of minimizing the discrepancy of ``source $\to$ target'' (\textit{i.e.,} $d(P_{s},P_{\rm inter})$) is easier than minimizing the discrepancy of ``target $\to$ intermediate'' (\textit{i.e.,} $d(P_{t},P_{\rm inter})$).
Thus, we force the minimization objective to penalize more on $d(P_{s},P_{\rm inter})$ than $d(P_{t},P_{\rm inter})$ by multiplying $d(P_{s},P_{\rm inter})$ by $a^{s}$ and multiplying $d(P_{t},P_{\rm inter})$ by $a^{t}$. By utilizing the mix ratios as the weighting coefficients in Eq.~(\ref{eq:loss_mix}), the alignment between the intermediate domains and the source and target domains can be adaptively conducted.

In a deep model for UDA re-ID, we consider to enforce the bridge loss (Eq. (\ref{eq:loss_mix})) on the feature and prediction space of the network to align the source and target by the intermediate domains. 
For the prediction space, we use the cross-entropy to measure the discrepancy between intermediate domains' prediction logits and other two extreme domains' (pseudo) labels (Eq. (\ref{eq:loss_mix_label})). For the feature space, we use the L2-norm to measure features' distance among domains (Eq. (\ref{eq:loss_mix_feat})). Our proposed two bridge losses are formulated as follows:
\begin{equation}
\small
\mathcal L_{\rm bridge}^{\varphi}=-\frac{1}{n}\sum_{i=1}^{n}\sum_{k\in \{s,t\}}a_{i}^{k}\cdot \left [y_{i}^{k}log(\varphi(f_{m}(G_{i}^{\rm inter})))\right ],
\label{eq:loss_mix_label}
\end{equation}
\begin{equation}
\small
\mathcal L_{\rm bridge}^{f}=\frac{1}{n}\sum_{i=1}^{n}\sum_{k\in \{s,t\}}a_{i}^{k}\cdot \left \|f_{m}(G^{k}_{i}) -f_{m}(G_{i}^{\rm inter}) \right \|_{2}.
\label{eq:loss_mix_feat}
\end{equation}
In Eq.~(\ref{eq:loss_mix_label}) (\ref{eq:loss_mix_feat}), we use $k$ to indicate the domain (source or target) and use $i$ to index the data in a mini-batch. $G^{k}_{i}$ is the $k$ domain's representation at the $m$-th stage.
$G_{i}^{\rm inter}$ is the intermediate domain's representation at the $m$-th hidden stage by mixing $G_{i}^{s}$ and $G_{i}^{t}$ as in Eq. (\ref{eq:Mix}).
The $f_{m}(\cdot)$ is the mapping from the $m$-th hidden stage to the features after GAP layer and $\varphi(\cdot)$ is the classifier.

In conclusion, we first utilize IDM to generate the intermediate domains by  mixing up the representations of the source and target (with two corresponding mix ratios ($a^{s}$ and $a^{t}$)) at the $m$-th hidden stage. Second, we propose the diversity loss $\mathcal{L}_{\rm div}$ by maximizing the diversity of the mix ratios, in order to generate intermediate domains bridging the source and target as diverse as possible. Third, we propose the bridge losses (\textit{i.e.,} $\mathcal L_{\rm bridge}^{\varphi}$ and $\mathcal L_{\rm bridge}^{f}$) to adaptively minimize the discrepancy of ``source $\to$ intermediate'' and the discrepancy of ``target $\to$ intermediate''.

\subsection{Mirrors Generation Module}
\label{sec:domain_mapping}

We recall that when IDM mixes a source-domain and a target-domain feature to obtain an intermediate-domain feature, it has an side-effect of mixing their identities (Section~\ref{sec:IDM-modelling}). Consequentially, the intermediate domains do not contain the original training identities. We argue that these original identities are also beneficial for domain alignment, because anchoring the identity allows focusing on the cross-domain variance and thus benefits the domain alignment.
To this end, Mirrors Generation Module (MGM) 1) maps the original source and target-domain identities into the intermediate domains and obtain their mirrors and 2) uses a cross-domain consistency loss between the mirrors and their original features for domain alignment.

\textbf{Mapping the features into the intermediate domains.}
Given a sample's feature map $G\in \mathbb{R}^{H\times W\times C}$ at the $l$-th hidden stage of the network, its feature statistics are denoted as the mean $\mu (G)\in \mathbb{R}^{C}$ and the standard deviation $\sigma(G)\in \mathbb{R}^{C}$, where each channel is calculated across spatial dimensions independently as follows:
\begin{equation}
\small
    \mu_{c} (G)=\frac{1}{H\times W}\sum_{h=1}^{H}\sum_{w=1}^{W}G_{hwc},
\end{equation}
\begin{equation}
\small
    \sigma_{c}(G)=\sqrt{\frac{1}{H\times W}\sum_{h=1}^{H}\sum_{w=1}^{W}(G_{hwc}-\mu_{c}(G))^{2}}.
\end{equation}

Motivated by recent studies on out-of-distribution \cite{zhou2021domain,nuriel2021permuted,tang2021crossnorm}, feature statistics (\textit{i.e.,} $\mu$, $\sigma$) are utilized to represent a sample's domain characteristics and the normalized content (\textit{i.e.,} $(G-\mu)/\sigma$) is utilized to represent a sample's identity-specific information.
For a sample's feature map $G^{a}$ from the domain ``$a$'' and a sample's feature map $G^{b}$ from the domain ``$b$'', we utilize AdaIN~\cite{huang2017arbitrary} to map $G^{a}$ into the domain ``$b$'' as follows:
\begin{equation}
\small
    G^{a\rightarrow b} = \mathrm{AdaIN}(G^{a},G^{b})=\sigma(G^{b})\left ( \frac{G^{a}-\mu(G^{a})}{\sigma(G^{a})}\right)+\mu(G^{b}).
\label{eq:adain}
\end{equation}
By utilizing $\mathrm{AdaIN}(G^{a},G^{b})$, we can easily characterize the sample's feature $G^{a}$ with the style of the domain ``$b$'' to get the mirror feature $G^{a \rightarrow b}$. Compared with $G^{a}$, $G^{a \rightarrow b}$ can represent the same person identity but different domain style.

Similar to  the formulation in Section~\ref{sec:IDM-module}, we can obtain source, target, and intermediate domains' feature maps $\left \{ (G_{i}^{s},G_{i}^{t},G_{i}^{\rm inter}) \right \}_{i=1}^{n}$ at the $l$-th hidden stage of the network. Then we utilize Eq. (\ref{eq:adain}) to characterize the source and target feature maps with intermediate domains' statistics as follows:
\begin{equation}
\small
\left\{\begin{matrix}
G^{s\rightarrow \rm inter}_{i}=\mathrm{AdaIN}(G^{s}_{i},G^{\rm inter}_{i})
\\ 
G^{t\rightarrow \rm inter}_{i}=\mathrm{AdaIN}(G^{t}_{i},G^{\rm inter}_{i})
\end{matrix}\right. .
\label{eq:adain2}
\end{equation}
With Eq. (\ref{eq:adain2}), we can obtain the source mirrors $\{G^{s \rightarrow  \rm inter}_{i}\}_{i=1}^{n}$ and target mirrors $\{G^{t \rightarrow \rm inter}_{i}\}_{i=1}^{n}$ characterized by intermediate domain styles. We denote the operation of Eq.~(\ref{eq:adain2}) as generating mirrors in MGM in Fig.~\ref{fig:pipeline}~(c).

\textbf{Cross-domain consistency.} MGM uses cross-domain consistency to make the original identity-discriminative information consistent across different domains. Specifically, MGM enforces a consistency loss between the prediction space $\varphi$ of samples and the $\varphi$ of their mirrors. For simplicity, we denote predictions of $n$ samples in a domain as $\{ \varphi_{i} \}_{i=1}^{n}$, representing an identity's distribution characteristic.

In the prediction space, we conduct the softmax function on the logits after the classifier to model the prediction distribution. Given the feature map $G$ at the $l$-th hidden stage of the network, and the (pseudo) label $y$, the prediction distribution is formulated as follows:
\begin{equation}
\small
    P(y|G)=\frac{\mathrm{exp}(\varphi_{y}(f_{l}(G))/\tau)}{\sum\nolimits_{i=1}^{C^{s}+C^{t}}\mathrm{exp}(\varphi_{i}(f_{l}(G))/\tau)},
\label{eq:prediction_logits}
\end{equation}
where $f_{l}(\cdot)$ denotes the part of the network mapping $G$ after the $l$-th hidden stage to the 2048-dim feature after the GAP layer, $\varphi_{i}$ denotes the logit of the classifier for class $i$, and $\tau>0$ is the temperature scaling parameter. Thus, we propose a class-wise consistency loss that enforces consistency on predictions of samples and their mirrors in intermediate domains. Given a sample's feature map $G^{a}$ from domain ``$a$'' and its mirror (sharing the same identity of $G^{a}$) $G^{a\rightarrow b}$ stylized by domain ``$b$'', their discrepancy in the prediction space is calculated as follows:
\begin{equation}
\small
\begin{split}
    D_{\varphi}(G^{a},G^{a\rightarrow b}) &= \mathrm{KL}\left ( P(y|G^{a}) || P(y|G^{a\rightarrow b}) \right ) \\
    &+ \mathrm{KL}\left ( P(y|G^{a\rightarrow b}) || P(y|G^{a}) \right ),
\end{split}
\label{eq:KL-prediction}
\end{equation}
where KL is the Kullback-Leibler (KL) divergence. With Eq.~(\ref{eq:KL-prediction}), the cross-domain consistency loss is formulated as follows:
\begin{equation}
\small
\mathcal{L}_{\mathrm{cons}}=\tau^{2}\cdot \left ( D_{\varphi}(G^{s},G^{s\rightarrow \rm inter})+D_{\varphi}(G^{t},G^{t\rightarrow \rm inter}) \right ).
\label{eq:loss-align-prediction}
\end{equation}
Following the knowledge distillation method \cite{hinton2015distilling}, we multiple the square of the temperature $\tau^{2}$. By minimizing Eq.~(\ref{eq:loss-align-prediction}), we can encourage the model to focus on minimizing the diverse domain variation during the domain alignment.

\begin{algorithm}[tp]
\scriptsize
\caption{The overall training procedure}
\label{alg:algorithm1}
\KwIn{
Source dataset $\{(x^{s}_{i},y^{s}_{i})\}$ and target dataset $\{x^{t}_{i}\}$;
}
\KwOut{
The trained backbone network $f(\cdot)$ and classifier $\varphi(\cdot)$;
}
    
Initialize the ImageNet-pretrained backbone network $f(\cdot)$;

Initialize the XBM memory as an empty queue $M$;

Plug our IDM and MGM at the $m$-th and $l$-th stage respectively. 

\For{$epoch=1$ to MaxEpochs}
{
Use $f(\cdot)$ to extract features $\{f^{t}_{i}\}$ for the target dataset $\{x^{t}_{i}\}$;

Assign pseudo labels $\{y^{t}_{i}\}$ for target samples $\{x^{t}_{i}\}$ by performing DBSCAN clustering on their features $\{f^{t}_{i}\}$;

\For{$iter=1$ to MaxIters}
{

Sample a mini-batch including $n$ source samples $\{(x^{s}_{i},y^{s}_{i})\}_{i=1}^{n}$ and $n$ target samples $\{(x^{t}_{i},y^{t}_{i})\}_{i=1}^{n}$;

Using IDM to obtain $f^{\rm inter}$ and $\varphi^{\rm inter}$;

Using MGM to obtain $\varphi^{s\rightarrow inter}$ and $\varphi^{t\rightarrow inter}$;

\If {using XBM}{

Enqueue($M$, $\{(f^{t},y^{t})\}$, $\{(f^{s},y^{s})\}$);

\If {M is full}{
Dequeue($M$);
}
Use all entries in $M$ for the hard negatives mining in the triplet loss in $\mathcal L_{\rm ReID}$ from Eq.~(\ref{eq:overall_loss});

}

Optimizing the network by the gradients of Eq. (\ref{eq:overall_loss});

}
}
\end{algorithm}

\textbf{Overall training for IDM++.}
Commonly, the ResNet-50 backbone contains five stages, where stage-0 is comprised of the first Conv, BN and Max Pooling layer and stage-1/2/3/4 correspond to the other four convolutional blocks.
We plug our IDM at the $m$-th stage of ResNet-50 (\textit{e.g.,} stage-0 as shown in Fig.~\ref{fig:pipeline}) to generate intermediate domains, and plug the MGM at the $l$-th stage (\textit{e.g.,} stage-3 in Fig.~\ref{fig:pipeline}). The overall training loss is as follows:
\begin{equation}
\small
    \mathcal L= \mathcal L_{\rm ReID}+\mu_{1}\cdot \mathcal L_{\rm bridge}^{\varphi}+\mu_{2}\cdot \mathcal L_{\rm bridge}^{f}+\mu_{3}\cdot \mathcal L_{\rm div} + \mu_{4} \cdot  \mathcal L_{\rm cons},
\label{eq:overall_loss}
\end{equation} 
where $\mathcal L_{\rm ReID} = (1-\mu_{1}) \cdot \mathcal L_{\rm cls} + \mathcal L_{\rm tri}$, and $\mu_{1}$, $\mu_{2}$, $\mu_{3}$, $\mu_{4}$ are the weights to balance losses. The training procedure is shown in Algorithm \ref{alg:algorithm1}.

\subsection{Extension to Domain Generalizable Person Re-ID}
\label{sec:extension_dg_reid}

Compared with UDA re-ID, domain generalizable re-ID (DG re-ID) is more challenging because target domain data can not be accessible during training. Different from UDA re-ID, the keynote of DG re-ID is to make the deep model robust to the variation of multiple source domains, so that the learned features can be well generalized to any unseen target domain. 
For example, the existing DG re-ID works~\cite{song2019generalizable,jin2020style,zhao2021heterogeneous} also call these features as domain-invariant features, or identity-relevant features. 

While our IDM++ is initially designed for UDA, we may apply it to DG scenario through slight modifications. Specifically, we merge multiple domains (without using domain labels) into a hybrid training domain. During training, each mini-batch contains features from multiple domains. 
For a mini-batch of $n$ samples, we first obtain their feature maps $\{ G_{i}^{s} \}_{i=1}^{n}$ at the $m$-th hidden stage and randomly shuffle them along the batch dimension to obtain out-of-order feature maps $\{ \widetilde{G_{i}^{s}} \}_{i=1}^{n}$. Next, we consider $\{ \widetilde{G_{i}^{s}} \}_{i=1}^{n}$ as pseudo target domains' feature maps and use IDM to generate intermediate domains' representations with Eq.~(\ref{eq:Mix}). When mapping source identities into intermediate domains with the MGM, we just use $\mathrm{AdaIN}(\cdot, \cdot)$ to change statistics values of the source-domain features with statistics values of the intermediate-domain features and obtain source mirrors $\{G^{s \rightarrow \rm inter}_{i}\}_{i=1}^{n}$. Next, we only enforce the consistency loss $\mathcal{L}_{\rm cons}$ between  $\{G^{s \rightarrow \rm inter}_{i}\}_{i=1}^{n}$ and $\{ G_{i}^{s} \}_{i=1}^{n}$. The learning objective of DG re-ID is the same as the objective of UDA re-ID (\textit{i.e.,} Eq.~(\ref{eq:overall_loss})). 

Through the above training procedure, IDM++ improves the robustness against domain variation. There are two main reasons for the improvement. First, IDM adds more source domains for training, because the synthetic intermediate domains may be viewed as extra source domains for training. Second, MGM further enforces consistency among multiple domains and promotes learning domain-invariant features. Therefore, IDM and MGM jointly helps IDM++ to enhance the feature robustness against domain variation, which benefits generalization towards novel unseen target domains.

\section{Experiments}\label{sec:experiments}
We introduce the datasets and evaluation protocols in Section~\ref{sec:datasets}. Section~\ref{sec:implementation} describes the implementation details of IDM++ for both UDA and DG re-ID, respectively. 
Section~\ref{sec:ablation} conducts ablation study to analyze the improvement of IDM++, as well as its two key components (\textit{i.e.,} IDM and MGM). 
Section~\ref{sec:parameter} analyzes the impact of some important hyper-parameters.
Section~\ref{sec:SOTA} compares our IDM++ with the state-of-the-art methods on twelve UDA re-ID benchmarks and two protocols of DG re-ID. 
Section~\ref{sec:investigation_on_mechanism} validates some key designs within IDM and MGM, while Section~\ref{sec:analysis} investigates the mechanism of IDM++ and reveals how the intermediate domains improve the domain alignment.

\subsection{Datasets and Evaluation Protocols}
\label{sec:datasets}
\textbf{Datasets.}
A total of six person re-ID datasets are used in our experiments, including Market-1501 \cite{zheng2015scalable}, DukeMTMC-reID \cite{ristani2016performance,zheng2017unlabeled}, CUHK03 \cite{li2014deepreid}, MSMT17 \cite{wei2018person}, PersonX \cite{sun2019dissecting}, and Unreal \cite{zhang2021unrealperson}. Among these datasets, PersonX and Unreal are synthetic datasets, and all the remaining datasets are from the real-world. We adopt CUHK03-NP \cite{zhong2017re} when testing on CUHK03.

\textbf{Evaluation Protocols.}
We use mean average precision (mAP) and Rank-1/5/10 (R1/5/10) of CMC to evaluate performances. In training, we do not use any additional information like temporal consistency in JVTC+ \cite{li2020joint}.
In testing, there are no post-processing techniques like re-ranking \cite{zhong2017re} or multi-query fusion \cite{zheng2015scalable}. We follow the mainstream evaluation protocols to evaluate our method in both UDA re-ID and DG re-ID tasks. 
1) UDA re-ID protocols: During training, we train on a labeled source dataset and an unlabeled target dataset. We use mAP and CMC to test the model on the target dataset. Specifically, we evaluate on two kinds of UDA re-ID tasks, \textit{i.e.,} real $\to$ real and synthetic $\to$ real, where the source datasets are real and synthetic respectively.
2) DG re-ID protocols: Following recent DG re-ID works \cite{zhao2021learning,dai2021generalizable}, we adopt the leave-one-out setting. Given a pool of datasets (Market-1501, DukeMTMC-reID, CUHK03, and MSMT17), we use a single dataset for testing and use all the other datasets for training. 
Specifically, we use two protocols, the ``DG-partial'' \cite{zhao2021learning} and the ``DG-full'' \cite{dai2021generalizable}. DG-partial uses only the training set while DG-full combines the training and testing set for training.

\subsection{Implementation Details}
\label{sec:implementation}
ResNet-50 \cite{he2016deep} pretrained on ImageNet is adopted as the backbone network. Following \cite{ge2020self}, domain-specific BNs \cite{chang2019domain} are also used in the backbone network to narrow domain gaps.
Following \cite{luo2019strong}, we resize the image size to 256$\times$128 and apply some common image augmentation techniques, including random flipping, random cropping, and random erasing \cite{zhong2020random}. For UDA re-ID, we perform DBSCAN \cite{ester1996density} clustering on the unlabeled target data to assign pseudo labels before each training epoch, which is consistent with  \cite{fu2019self,song2020unsupervised,ge2020self}.
The mini-batch size is 128, including 64 source images of 16 identities and 64 target images of 16 pseudo identities. We totally train 50 epochs and each epoch contains 400 iterations. The initial learning rate is set as $3.5\times 10^{-4}$ which will be divided by 10 at the 20th and 40th epoch respectively. The Adam optimizer with weight decay $5\times 10^{-4}$ and momentum 0.9 is adopted in our training. We implement Strong Baseline of UDA re-ID with XBM \cite{wang2020cross} to mine more hard negatives for the triplet loss.
For DG re-ID, the training setting follows BoT \cite{luo2019strong}. Furthermore, in order to prevent learning identity-discriminative features from getting into collapse solutions when enforcing the consistency loss in DG re-ID, we use the twin-branch network structure (duplicating the branch from the $l-$th stage to the classifier), and feed the original samples and mirrors into different branches. Similar to BYOL~\cite{byol}, we enforce the consistency loss on the outputs of the twin-branches.
If not specified, the loss weights $\mu_{1}, \mu_{2}, \mu_{3}, \mu_{4}$ are set as 0.7, 0.1, 1.0, 1.0 respectively, and the temperature $\tau$ is set as 0.5 in all our experiments.
In the IDM module, the FC1 layer is parameterized by $W_{1}\in \mathbb{R}^{c\times 2c}$  and MLP is composed of two fully connected layers (followed by batch normalization) which are parameterized by $W_{2}\in \mathbb{R}^{(c/r) \times c}$ and $W_{3}\in \mathbb{R}^{2 \times (c/r)}$ respectively, where $c$ is the representations' channel number after the $m$-th stage and $r$ is the reduction ratio. If not specified, we plug the IDM module at the stage-0 of ResNet-50 and set $r$ as 2, and plug the MGM at the stage-3. The IDM and MGM modules are only used in training and will be discarded in testing. Our method is implemented with Pytorch, and four Nvidia Tesla V100 GPUs are used for training and only one GPU is used for testing.

\begin{table*}[htp]
\begin{center}
\caption{Ablation studies on different components of our method. Baseline1 (Naive Baseline): only using $\mathcal L_{\rm ReID}$ to train the source and target domains jointly. Baseline2 (Strong Baseline): Baseline1 + XBM \cite{wang2020cross}.}
\scriptsize
\label{tab:ablation}
\begin{tabular}{l|ccc|c|cc|cc}
\hline
\multirow{2}{*}{Method} & \multicolumn{3}{c|}{IDM} & MGM & \multicolumn{2}{c|}{DukeMTMC$\to$Market-1501} & \multicolumn{2}{c}{Market-1501$\to$DukeMTMC} \\ \cline{2-9} 
 & $\mathcal L_{\rm bridge}^{\varphi}$ & $\mathcal L_{\rm bridge}^{f}$ & $\mathcal L_{\rm div}$ & $\mathcal L_{\rm cons}$ & mAP & R1 & mAP & R1 \\ \hline
Oracle &  &  &  &  &83.9  &93.2  &75.0  &86.1  \\
Baseline1 &  &  &  &  &77.0  &90.6  &63.4  &78.4  \\ \hline
\multirow{4}{*}{IDM} &  & \checkmark & \checkmark &  &79.4 &91.5 &66.2 &79.8  \\
 & \checkmark &  & \checkmark &  &79.2 &91.2 &65.8 &80.4  \\
 & \checkmark & \checkmark &  &  &78.9 &91.1 &64.8 &79.3  \\
 & \checkmark & \checkmark & \checkmark &  &81.9 &92.4 &68.6 &82.3  \\ \hline
\multirow{4}{*}{IDM++} &  & \checkmark & \checkmark & \checkmark &81.2 &92.3 &68.5 &81.8 \\
 & \checkmark &  & \checkmark & \checkmark &81.4 &91.8 &69.5 &82.1 \\
 & \checkmark & \checkmark &  & \checkmark &80.9 &92.0 &69.9 &82.4 \\
 & \checkmark & \checkmark & \checkmark & \checkmark &82.4 &92.7 &71.6 &83.4 \\ \hline \hline
Oracle &  &  &  &  &86.9 &94.8 &78.1 &88.3  \\
Baseline2 &  &  &  &  &79.1  &91.2  &65.8 &80.1  \\ \hline
\multirow{4}{*}{IDM} &  & \checkmark & \checkmark &  &80.7 &92.0 &67.3 &81.8  \\
 & \checkmark &  & \checkmark &  &81.6 &92.2 &69.0 &82.0  \\
 & \checkmark & \checkmark &  &  &79.7 &91.6 &67.8 &81.6  \\
 & \checkmark & \checkmark & \checkmark &  &82.8 &93.2 &70.5 &83.6  \\ \hline
\multirow{4}{*}{IDM++} &  & \checkmark & \checkmark & \checkmark &83.4 &93.1 &70.0 &83.0 \\
 & \checkmark &  & \checkmark & \checkmark &84.3 &93.4 &72.2 &83.9 \\
 & \checkmark & \checkmark &  & \checkmark &83.8 &93.3 &72.5 &84.4 \\
 & \checkmark & \checkmark & \checkmark & \checkmark &85.3 &94.2 &73.2 &85.5 \\ 
\hline
\end{tabular}
\end{center}
\vspace{-1.5em}
\end{table*}

\begin{figure*}[htp]
		\centering
		\footnotesize
		\setlength\tabcolsep{1mm}
		\renewcommand\arraystretch{0.1}
		\begin{tabular}{ccccc}
			\includegraphics[width=0.18\linewidth]{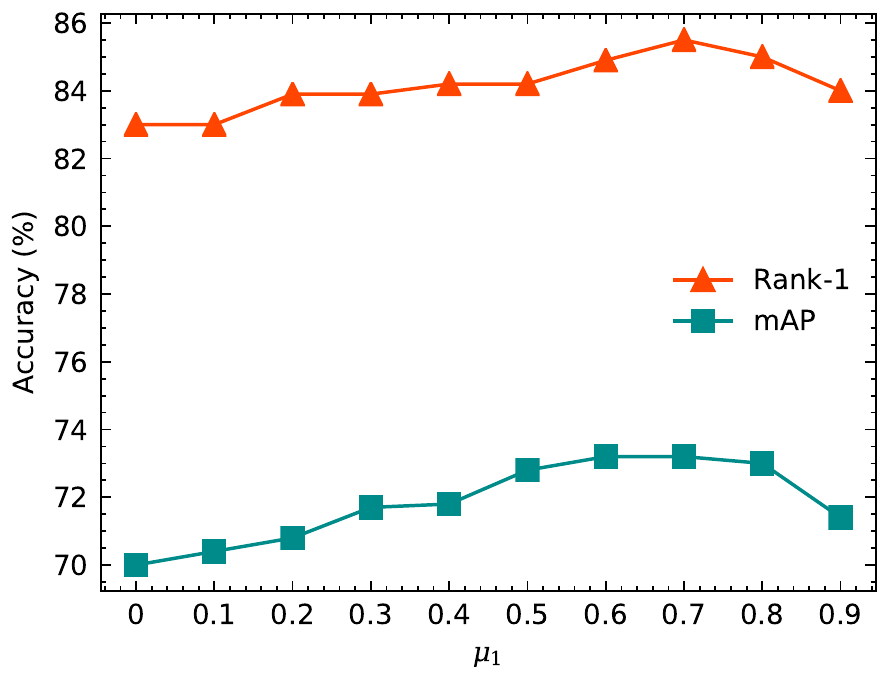} &
			\includegraphics[width=0.18\linewidth]{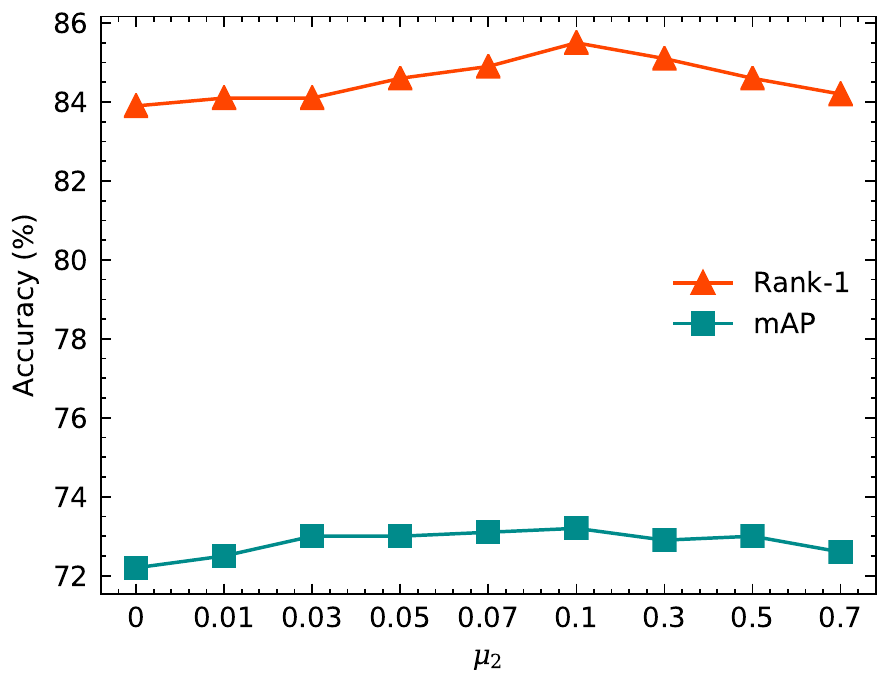} &
			\includegraphics[width=0.18\linewidth]{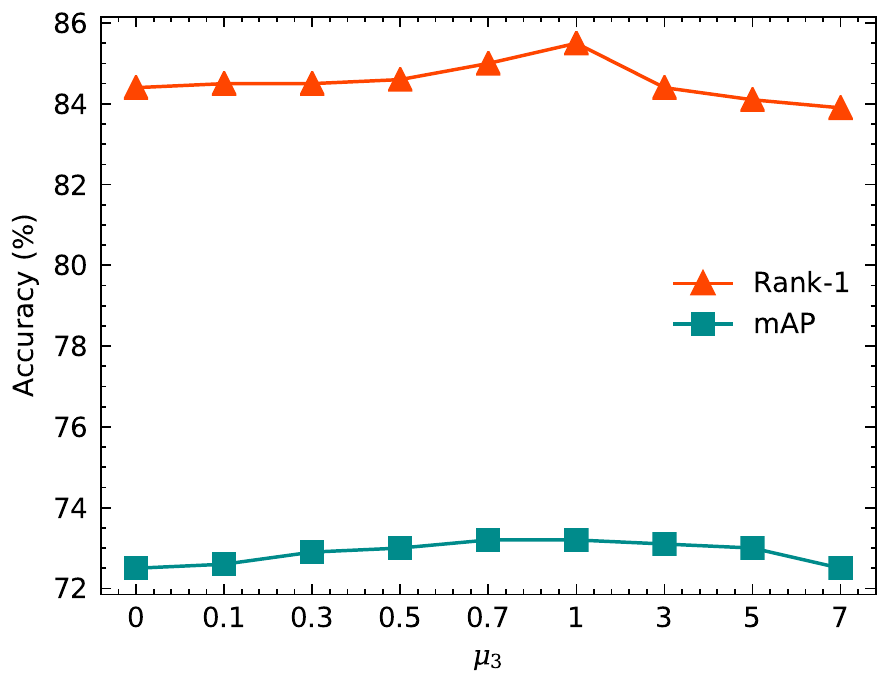} &
			\includegraphics[width=0.18\linewidth]{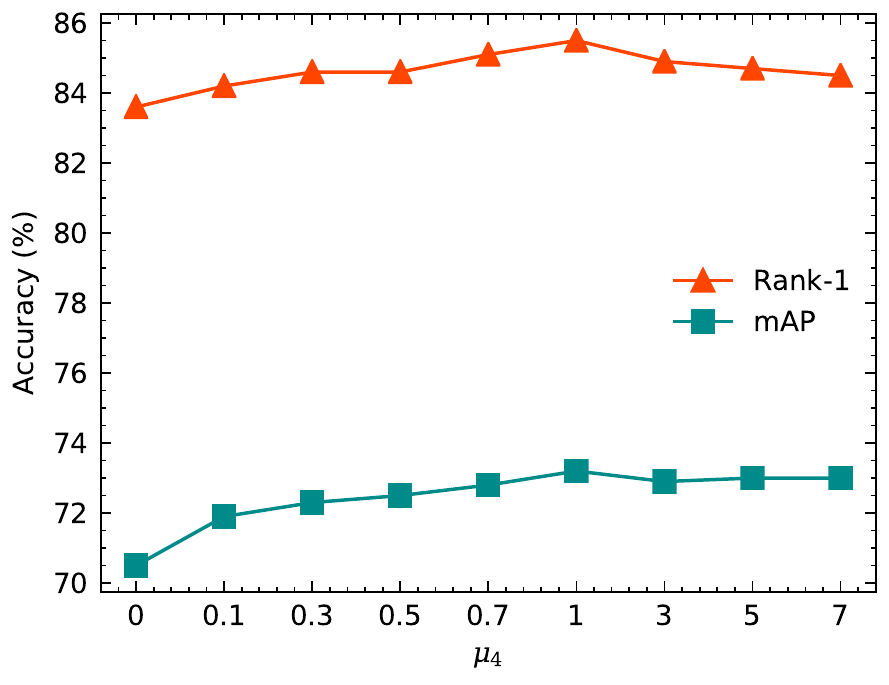} &
			\includegraphics[width=0.18\linewidth]{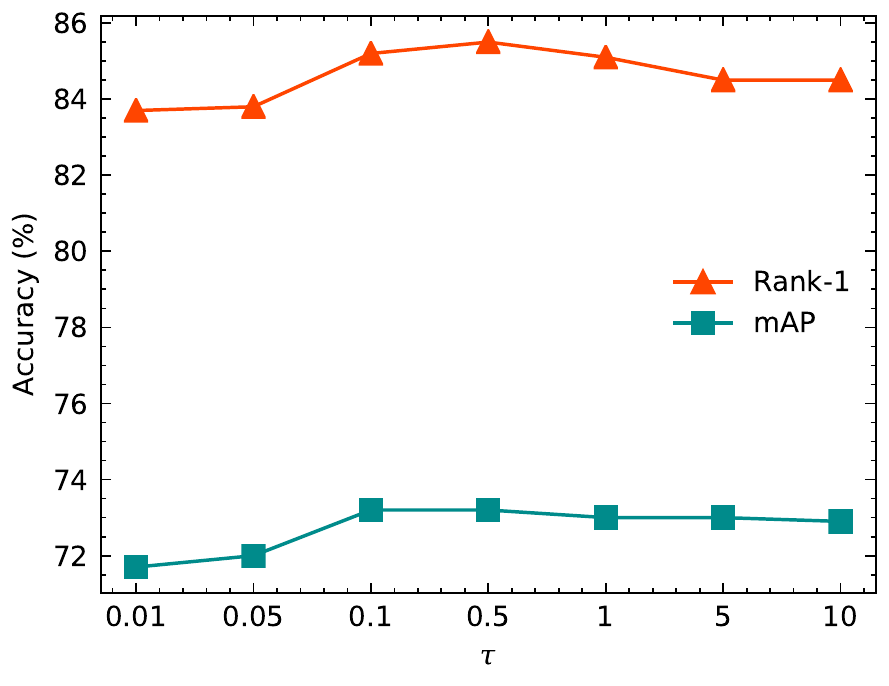} \\
			~\\
    			(a) Loss weight $\mu_{1}$ & (b) Loss weight $\mu_{2}$ & (c) Loss weight $\mu_{3}$ & (d) Loss weight $\mu_{4}$ & (e) Temperature $\tau$
		\end{tabular}
		\caption{Analysis of different values of hyper-parameters: $\mu_{1}$, $\mu_{2}$, $\mu_{3}$, $\mu_{4}$, and $\tau$ on the target domain when transferring from Market-1501 to DukeMTMC-reID.}
		\label{fig:params}
\end{figure*}

\subsection{Ablation Study}
\label{sec:ablation}
This section evaluates the effectiveness of each component in IDM++ and the results are provided in Table~\ref{tab:ablation}. The ``Oracle'' method in Table~\ref{tab:ablation} uses the ground-truth label for the target (and source) domain, marking the upper bound of UDA re-ID accuracy.

\textbf{IDM improves the baseline.} IDM is inserted in the hidden stage of the network to generate intermediate domains' features that can be utilized to better align the source and target domains. In IDM, there are two functions: 1) using the diversity loss (\textit{i.e.,} $\mathcal L_{\rm div}$) to generate dense intermediate domains as many as possible and 2) using the bridge losses (\textit{i.e.,} $\mathcal L_{\rm bridge}^{\phi}$ and $\mathcal L_{\rm bridge}^{f}$) to dynamically minimize the discrepancy between the intermediate domains and the source and target domains. We evaluate the effectiveness of both functions in Table \ref{tab:ablation}.
(1) The effectiveness of $\mathcal L_{\rm div}$: ``Baseline2 + IDM w/o $\mathcal L_{\rm div}$'' degenerates much when comparing with the full method. The Rank-1 of ``Baseline2 + IDM++ w/o  $\mathcal L_{\rm div}$'' is 1.1\% lower than the full method on Market $\to$ Duke. 
(2) The effectiveness of $\mathcal L_{\rm bridge}^{\phi}$ and $\mathcal L_{\rm bridge}^{f}$: Taking Market$\to$Duke as an example, mAP/R1 of ``Baseline2 + IDM++ w/o $\mathcal L_{\rm bridge}^{\phi}$'' is 3.2\%/2.5\% lower than ``Baseline2 + IDM++(full)''.
The mAP/R1 of ``Baseline2 + IDM++ w/o $\mathcal L_{\rm bridge}^{f}$'' is 1.0\%/1.6\% lower than ``Baseline2 + IDM++(full)''. Compared with $\mathcal L_{\rm bridge}^{f}$, $\mathcal L_{\rm bridge}^{\phi}$ is more important based on Baseline2. The reason may be that Baseline2 uses XBM~\cite{wang2020cross} to mine harder negatives in the feature space, which will affect the effectiveness of~$\mathcal L_{\rm bridge}^{f}$.
The above analyses show that all the functions in the IDM are important for learning appropriate intermediate domains that can help to bridge the source and target domains.

\textbf{MGM achieves further improvement based on IDM.} Compared with our previous work (IDM \cite{dai2021idm}), this paper proposes a novel MGM to map source / target features into intermediate domains, which can help learn better identity-discriminative representations across domains. We first use the MGM to map the source and target domains' feature maps to intermediate domains at the $l$-th (if not mentioned, we set $l$ as 3) stage and then we use the class-wise domain consistency loss $\mathcal L_{\rm cons}$ to enforce the consistency on predictions of source and target samples together with their mirrors. In Table \ref{tab:ablation}, we evaluate the effectiveness of the newly proposed MGM compared with our previous IDM. Taking Market$\to$Duke as an example, mAP/R1 of ``Baseline2 + IDM++(full)'' is 2.7\%/1.9\% higher than ``Baseline2 + IDM(full)''.

\textbf{IDM++ integrating IDM and MGM brings general improvement on multiple baselines.} We propose two baseline methods including Naive Baseline (Baseline1) and Strong Baseline (Baseline2). Naive Baseline means only using $\mathcal L_{\rm ReID}$ to train the source and target data in joint manner as shown in Fig.~\ref{fig:pipeline}.
Compared with Naive Baseline, Strong Baseline uses XBM \cite{wang2020cross} to mine more hard negatives for the triplet loss, which is a variant of the memory bank \cite{xiao2017joint,wu2018unsupervised} and easy to implement. 
Similar to those UDA re-ID methods \cite{zhong2020learning,ge2020self} using the memory bank, we set the memory bank size as the number of all the training data.
As shown in Table \ref{tab:ablation}, no matter which baseline we use, our methods can obviously outperform the baseline methods by a large margin. On Market$\to$Duke, 
mAP/R1 of ``Baseline2 + our IDM++ (full)'' is 7.4\%/5.4\% higher than Baseline2, and mAP/R1 of ``Baseline1 + our IDM++ (full)'' is 8.2\% and 5.0\% higher than Baseline1.
Because of many state-of-the-arts methods \cite{ge2020self,zheng2020exploiting,li2020joint,zheng2021group} use the memory bank to improve the performance on the target domain, we use Strong Baseline to implement our IDM for fairly comparing with them.

\subsection{Parameter Analysis}
\label{sec:parameter}
In Fig.~\ref{fig:params}, we provide the visualization on the sensitivity of hyper-parameters. Specifically, we evaluate the loss weights: $\mu_{1}$, $\mu_{2}$, $\mu_{3}$, and $\mu_{4}$ in Eq. (\ref{eq:overall_loss}), and the temperature value $\tau$ in Eq. (\ref{eq:prediction_logits}). We conduct the experiments on Market-1501$\to$DukeMTMC-reID and evaluate the mAP and Rank-1 on the target domain: DukeMTMC-reID. As shown in Fig.~\ref{fig:params}, these hyper-parameters are not very sensitive. The performance achieves the best when we set $\mu_{1}$, $\mu_{2}$, $\mu_{3}$, $\mu_{4}$, and $\tau$ as 0.7, 0.1, 1.0, 1.0, and 0.5 respectively. If not specified, we utilize the above setting of hyper-parameters in all other experiments in this paper.

\begin{table*}[htp]
\caption{Comparison with the state-of-the-art UDA re-ID methods on real $\to$ real tasks. The best results are highlighted with bold and the second best results are marked with underline. 
}
\scriptsize
\label{tab:SOTA-real2real}
\begin{center}
\begin{tabular}{l|c|p{1cm}<{\centering}p{1cm}<{\centering}p{1cm}<{\centering}p{1cm}<{\centering}|p{1cm}<{\centering}p{1cm}<{\centering}p{1cm}<{\centering}p{1cm}<{\centering}}
\hline
\multirow{2}{*}{Methods} & \multirow{2}{*}{Reference} & \multicolumn{4}{c|}{DukeMTMC-ReID $\to$ Market-1501} & \multicolumn{4}{c}{Market-1501 $\rightarrow$ DukeMTMC-ReID} \\ \cline{3-10} 
                         &                            & mAP      & R1       & R5       & R10     & mAP      & R1       & R5      & R10     \\ \hline 
PUL \cite{fan2018unsupervised}                     & TOMM 2018                  & 20.5     & 45.5     & 60.7     & 66.7    & 16.4     & 30.0     & 43.4    & 48.5    \\
TJ-AIDL \cite{wang2018transferable}                 & CVPR 2018                  & 26.5     & 58.2     & 74.8     & 81.1    & 23.0     & 44.3     & 59.6    & 65.0    \\
SPGAN+LMP \cite{deng2018image}                & CVPR 2018                  & 26.7     & 57.7     & 75.8     & 82.4    & 26.2     & 46.4     & 62.3    & 68.0    \\
HHL \cite{zhong2018generalizing}                     & ECCV 2018                  & 31.4     & 62.2     & 78.8     & 84.0    & 27.2     & 46.9     & 61.0    & 66.7    \\
ECN \cite{zhong2019invariance}                     & CVPR 2019                  & 43.0     & 75.1     & 87.6     & 91.6    & 40.4     & 63.3     & 75.8    & 80.4    \\
PDA-Net \cite{li2019cross}                 & ICCV 2019                  & 47.6     & 75.2     & 86.3     & 90.2    & 45.1     & 63.2     & 77.0    & 82.5    \\
PCB-PAST \cite{zhang2019self}                & ICCV 2019                  & 54.6     & 78.4     & -        & -       & 54.3     & 72.4     & -       & -       \\
SSG \cite{fu2019self}                     & ICCV 2019                  & 58.3     & 80.0     & 90.0     & 92.4    & 53.4     & 73.0     & 80.6    & 83.2    \\
MMCL \cite{wang2020unsupervised}                 & CVPR 2020                 & 60.4     & 84.4     & 92.8     & 95.0    & 51.4     & 72.4     & 82.9    & 85.0    \\
ECN-GPP \cite{zhong2020learning}                 & TPAMI 2020                 & 63.8     & 84.1     & 92.8     & 95.4    & 54.4     & 74.0     & 83.7    & 87.4    \\
JVTC+ \cite{li2020joint} &ECCV 2020 &67.2 &86.8 &95.2 &97.1 &66.5 &80.4 &89.9 &92.2  \\
AD-Cluster \cite{zhai2020ad}              & CVPR 2020                  & 68.3     & 86.7     & 94.4     & 96.5    & 54.1     & 72.6     & 82.5    & 85.5    \\
MMT \cite{ge2020mutual}                     & ICLR 2020                  & 71.2     &87.7     &94.9     &96.9    &65.1     &78.0     &88.8    &92.5    \\ 
CAIL \cite{luo2020generalizing} &ECCV 2020 &71.5 &88.1 &94.4 &96.2 &65.2 &79.5 &88.3 &91.4 \\
NRMT \cite{zhao2020unsupervised} &ECCV 2020  &71.7 &87.8 &94.6 &96.5 &62.2 &77.8 &86.9 &89.5 \\
MEB-Net \cite{zhai2020multiple} &ECCV 2020 &76.0 &89.9 &96.0 &97.5 &66.1 &79.6 &88.3 &92.2 \\ 
SpCL \cite{ge2020self} &NeurIPS 2020 &76.7 &90.3 &96.2 &97.7 &68.8 &82.9 &90.1 &92.5 \\ 
Dual-Refinement \cite{dai2020dual} &TIP 2021 &78.0 &90.9 &96.4 &97.7 &67.7 &82.1 &90.1 &92.5 \\
UNRN \cite{zheng2020exploiting} &AAAI 2021 &78.1 &91.9 &96.1 &97.8 &69.1 &82.0 &90.7 &93.5    \\ 
GLT \cite{zheng2021group} &CVPR 2021 &79.5 &92.2 &96.5 &97.8 &69.2 &82.0 &90.2 &92.8    \\
HCD \cite{zheng2021online} &ICCV 2021 &80.0 &91.5 &- &- &70.1 &82.2 &- &- \\
CCL \cite{isobe2021towards} &ICCV 2021 &\underline{83.4} &\textbf{94.2} &- &- &\underline{70.8} &83.5 &- &-  \\ \hline
Our IDM \cite{dai2021idm} & ICCV 2021  &82.8 &\underline{93.2} &\underline{97.5} &\underline{98.1} &70.5 &\underline{83.6} &\underline{91.5} &\underline{93.7}  \\ 
Our IDM++ & This paper  &\textbf{85.3} &\textbf{94.2} &\textbf{97.7} &\textbf{98.5} &\textbf{73.2} &\textbf{85.5} &\textbf{92.1} &\textbf{94.3} \\
\hline \hline

\multirow{2}{*}{Methods} & \multirow{2}{*}{Reference} & \multicolumn{4}{c|}{MSMT17 $\to$ Market-1501} & \multicolumn{4}{c}{MSMT17 $\to$ DukeMTMC-reID} \\ \cline{3-10} 
 &  & mAP & R1 & R5 & R10 & mAP & R1 & R5 & R10 \\ \hline
CASCL \cite{wu2019unsupervised} &ICCV 2019 &35.5 &65.4 &80.6 &86.2 &37.8 &59.3 &73.2 &77.8 \\
MAR \cite{yu2019unsupervised} &CVPR 2019 &40.0 &67.7 &81.9 &87.3 &48.0 &67.1 &79.8 &84.2 \\ 
PAUL \cite{yang2019patch} &CVPR 2019 &40.1 &68.5 &82.4 &87.4 &53.2 &72.0 &82.7 &86.0 \\ 
DG-Net++ \cite{zou2020joint} &ECCV 2020 &64.6 &83.1 &91.5 &94.3 &58.2 &75.2 &73.6 &86.9 \\  
D-MMD \cite{mekhazni2020unsupervised} &ECCV 2020 &50.8 &72.8 &88.1 &92.3 &51.6 &68.8 &82.6 &87.1 \\ 
MMT-dbscan \cite{ge2020mutual} &ICLR 2020 &75.6 &89.3 &95.8 &97.5 &63.3 &77.4 &88.4 &91.7 \\ 
SpCL \cite{ge2020self} &NeurIPS 2020 &77.5 &89.7 &96.1 &97.6 &69.3 &82.9 &91.0 &93.0 \\
HCD \cite{zheng2021online} &ICCV 2021 &80.2 &91.4 &- &- &71.2 &83.1 &- &- \\ \hline
Our IDM \cite{dai2021idm} & ICCV 2021 &\underline{82.1} &\underline{92.4} &\underline{97.5} &\underline{98.4} &\underline{71.9} &\underline{83.6} &\underline{91.5} &\underline{93.4} \\
Our IDM++ & This paper &\textbf{85.2} &\textbf{94.1} &\textbf{97.7} &\textbf{98.5} &\textbf{73.6} &\textbf{84.6} &\textbf{92.2} &\textbf{94.1} \\ \hline \hline

\multirow{2}{*}{Methods} & \multirow{2}{*}{Reference} & \multicolumn{4}{c|}{Market-1501 $\to$ MSMT17} & \multicolumn{4}{c}{DukeMTMC-reID $\to$ MSMT17} \\ \cline{3-10} 
 &  & mAP & R1 & R5 & R10 & mAP & R1 & R5 & R10 \\ \hline
ECN \cite{zhong2019invariance} & CVPR 2019 & 8.5 & 25.3 & 36.3 & 42.1 & 10.2 & 30.2 & 41.5 & 46.8 \\
SSG \cite{fu2019self} & ICCV 2019 & 13.2 & 31.6 & - & 49.6 & 13.3 & 32.2 & - & 51.2 \\
ECN-GPP \cite{zhong2020learning} & TPAMI 2020 & 15.2 & 40.4 & 53.1 & 58.7 & 16.0 & 42.5 & 55.9 & 61.5 \\
MMCL \cite{wang2020unsupervised} & CVPR 2020 & 15.1 & 40.8 & 51.8 & 56.7 & 16.2 & 43.6 & 54.3 & 58.9 \\
NRMT \cite{zhao2020unsupervised} &ECCV 2020 &19.8 &43.7 &56.5 &62.2 &20.6 &45.2 &57.8 &63.3 \\
CAIL \cite{luo2020generalizing} &ECCV 2020 &20.4 &43.7 &56.1 &61.9 &24.3 &51.7 &64.0 &68.9 \\
MMT \cite{ge2020mutual} & ICLR 2020 & 22.9 & 49.2 & 63.1 & 68.8 & 23.3 & 50.1 & 63.9 & 69.8 \\
JVTC+ \cite{li2020joint} &ECCV 2020 &25.1 &48.6 &65.3 &68.2 &27.5 &52.9 &70.5 &75.9 \\
SpCL \cite{ge2020self} &NeurIPS 2020 &26.8 &53.7 &65.0 &69.8 &26.5 &53.1 &65.8 &70.5 \\
Dual-Refinement \cite{dai2020dual} &TIP 2021 &25.1 &53.3 &66.1 &71.5 &26.9 &55.0 &68.4 &73.2 \\
UNRN \cite{zheng2020exploiting} & AAAI 2021 & 25.3 & 52.4 & 64.7 &69.7 & 26.2 & 54.9 & 67.3 & 70.6 \\ 
GLT \cite{zheng2021group} &CVPR 2021 &26.5  &56.6  &67.5  &72.0  &27.7  &59.5  &70.1  &74.2  \\
HCD \cite{zheng2021online} &ICCV 2021 &28.4 &54.9 &- &- &29.3 &56.1 &- &- \\
CCL \cite{isobe2021towards} &ICCV 2021 &\underline{35.8} &\underline{65.8} &- &- &\underline{36.3} &\underline{66.6} &- &- \\ \hline
Our IDM \cite{dai2021idm} & ICCV 2021 &33.5 &61.3 &\underline{73.9} &\underline{78.4} &35.4 &63.6 &\underline{75.5} &\underline{80.2} \\ 
Our IDM++ & This paper &\textbf{40.2} &\textbf{69.9} &\textbf{80.2} &\textbf{83.7} &\textbf{40.5} &\textbf{69.5} &\textbf{80.3} &\textbf{84.0} \\ \hline
\end{tabular}
\end{center}
\vspace{-1.5em}
\end{table*}

\subsection{Comparison with the State-of-the-arts}
\label{sec:SOTA}
The existing state-of-the-art (SOTA) UDA re-ID works commonly evaluate the performance on six real $\to$ real tasks \cite{fu2019self,wu2019unsupervised,zhong2020learning}. Recently, more challenging synthetic $\to$ real tasks \cite{ge2020self,zhang2021unrealperson} are proposed, where they use the synthetic dataset PersonX \cite{sun2019dissecting} or Unreal \cite{zhang2021unrealperson} as the source domain and test on other three real re-ID datasets.
Besides, the existing SOTA DG re-ID works usually evaluate the performance under two challenging ``leave-one-out'' protocols \cite{zhao2021learning,dai2021generalizable}, \textit{i.e.,} ``DG-partia'' and ``DG-full''. In the ``DG-partial'' protocol \cite{zhao2021learning}, the train-sets of three datasets among the four datasets are used for training and the test-set of the remaining dataset is used for testing. In the ``DG-full'' protocol \cite{dai2021generalizable}, all data (including train-sets and test-sets) of three datasets among the four datasets are used for training and the test-set of the remaining dataset is used for testing. To compare with the SOTAs fairly, we also take ResNet-50 as the backbone. All the results in Table \ref{tab:SOTA-real2real},~\ref{tab:SOTA-synthetic2real} show that our method can outperform the UDA re-ID SOTAs by a large margin, and the results in Table \ref{tab:SOTA-DG-reid} show that our method is superior to the DG re-ID SOTAs significantly.

\textbf{Comparisons on real $\to$ real UDA re-ID tasks.}
The existing UDA re-ID methods evaluated on real $\to$ real UDA tasks can be mainly divided into three categories based on their training schemes. (1) GAN transferring methods include PTGAN \cite{wei2018person}, SPGAN+LMP \cite{deng2018image}, \cite{zhong2018generalizing}, and PDA-Net \cite{li2019cross}. 
(2) Fine-tuning methods include PUL \cite{fan2018unsupervised}, PCB-PAST \cite{zhang2019self}, SSG \cite{fu2019self}, AD-Cluster \cite{zhai2020ad}, MMT \cite{ge2020mutual}, NRMT \cite{zhao2020unsupervised}, MEB-Net \cite{zhai2020multiple}, Dual-Refinement \cite{dai2020dual}, UNRN \cite{zheng2020exploiting}, and GLT \cite{zheng2021group}. 
(3) Joint training methods commonly use the memory bank \cite{xiao2017joint}, including ECN \cite{zhong2019invariance}, MMCL \cite{wang2020unsupervised}, ECN-GPP \cite{zhong2020learning}, JVTC+ \cite{li2020joint}, SpCL \cite{ge2020self}, HCD \cite{zheng2021online}, CCL \cite{isobe2021towards}. However, all these methods neglect the significance of intermediate domains, which can smoothly bridge the domain adaptation between the source and target domains to better transfer the source knowledge to the target domain. In our previous work IDM~\cite{dai2021idm}, we propose an IDM module to generate appropriate intermediate domains to better improve the performance of UDA re-ID. As an extension of IDM~\cite{dai2021idm}, we further propose IDM++ by integrating MGM and IDM to focus on the cross-domain variation when aligning the source and target domains.
As shown in Table \ref{tab:SOTA-real2real}, our method can outperform the second best UDA re-ID methods by a large margin on all these benchmarks. Especially when taking the most challenging MSMT17 as the target domain, our method outperforms the SOTA method CCL \cite{isobe2021towards} by 4.4\% mAP on Market-1501 $\to$ MSMT17 and 4.2\% mAP on DukeMTMC-reID $\to$ MSMT17. 
Adding MGM can significantly improve IDM, \textit{e.g.,} the performance gain is up to 6.7\% mAP and 8.6\% Rank-1 on Market-1501 $\to$ MSMT17.

\begin{table*}[htp]
\caption{Comparison with the state-of-the-art  UDA person re-ID methods on synthetic $\rightarrow$ real tasks. The best results are highlighted with bold and the second best results are marked with underline.
}
\label{tab:SOTA-synthetic2real}
\scriptsize
\begin{center}
\begin{tabular}{l|c|cccc|cccc|cccc}
\hline
\multirow{2}{*}{Methods} & \multirow{2}{*}{Reference} & \multicolumn{4}{c|}{PersonX $\to$ Market-1501} & \multicolumn{4}{c|}{PersonX $\to$ DukeMTMC-reID} & \multicolumn{4}{c}{PersonX $\to$ MSMT17} \\ \cline{3-14} 
 &  & mAP & R1 & R5 & R10 & mAP & R1 & R5 & R10 & mAP & R1 & R5 & R10 \\ \hline
MMT \cite{ge2020mutual} & ICLR 2020 &71.0 &86.5 &94.8 &97.0 &60.1 &74.3 &86.5 &90.5  &17.7  &39.1  &52.6  &58.5  \\
SpCL \cite{ge2020self} & NeurIPS 2020 &73.8 &88.0 &95.3 &96.9 &67.2 &81.8 &90.2 &92.6 &22.7 &47.7 &60.0 &65.5  \\
CCL \cite{isobe2021towards} &ICCV 2021 &79.6 &\underline{92.5} &- &- &- &- &- &- &28.9 &53.2 &- &- \\ \hline
Our IDM \cite{dai2021idm} & ICCV 2021 &\underline{81.3} &92.0 &\underline{97.4} &\underline{98.2} &\underline{68.5} &\underline{82.6} &\underline{91.2} &\underline{93.4} &\underline{30.3} &\underline{58.4} &\underline{70.7} &\underline{75.5}  \\ 
Our IDM++ & This paper &\textbf{84.3} &\textbf{93.7} &\textbf{97.5} &\textbf{98.4} &\textbf{71.5} &\textbf{84.1} &\textbf{91.9} &\textbf{94.0} &\textbf{37.7} &\textbf{68.3} &\textbf{79.1} &\textbf{82.8}  \\ \hline \hline
\multirow{2}{*}{Methods} & \multirow{2}{*}{Reference} & \multicolumn{4}{c|}{Unreal $\to$ Market-1501} & \multicolumn{4}{c|}{Unreal $\to$ DukeMTMC-reID} & \multicolumn{4}{c}{Unreal $\to$ MSMT17} \\ \cline{3-14} 
 &  & mAP & R1 & R5 & R10 & mAP & R1 & R5 & R10 & mAP & R1 & R5 & R10 \\ \hline
JVTC \cite{li2020joint} & ECCV 2020 &78.3 &90.8 &- &- &66.1 &81.2 &- &- &25.0 &53.7 &- &-  \\ \hline
Our IDM \cite{dai2021idm} &ICCV 2021 &\underline{83.2} &\underline{92.8} &\underline{97.3} &\underline{98.2} &\underline{72.4} &\underline{84.6} &\underline{92.0} &\underline{94.0} &\underline{38.3} &\underline{67.3} &\underline{78.4} &\underline{82.6} \\
Our IDM++ &This paper &\textbf{85.2} &\textbf{93.6} &\textbf{97.5} &\textbf{98.3} &\textbf{73.2} &\textbf{85.0} &\textbf{92.4} &\textbf{94.6} &\textbf{41.5} &\textbf{71.0} &\textbf{81.2} &\textbf{84.7} \\ \hline
\end{tabular}
\end{center}
\end{table*}

\begin{table*}[htp]
\caption{Comparison with the state-of-the-art DG person re-ID methods under the protocols of DG-partial and DG-full. The best results are highlighted with bold and the second best results are marked with underline.
}
\scriptsize
\label{tab:SOTA-DG-reid}
\begin{center}
\begin{tabular}{lcccp{1cm}<{\centering}p{1cm}<{\centering}cccp{1cm}<{\centering}p{1cm}<{\centering}}
\hline
\multicolumn{11}{c}{DG-partial} \\ \hline
\multicolumn{1}{l|}{\multirow{2}{*}{Method}} & \multicolumn{1}{c|}{\multirow{2}{*}{Source}} & \multicolumn{1}{c|}{\multirow{2}{*}{IDs}} & \multicolumn{1}{c|}{\multirow{2}{*}{Images}} & \multicolumn{2}{c|}{Target: Market-1501} & \multicolumn{1}{c|}{\multirow{2}{*}{Source}} & \multicolumn{1}{c|}{\multirow{2}{*}{IDs}} & \multicolumn{1}{c|}{\multirow{2}{*}{Images}} & \multicolumn{2}{c}{Target: DukeMTMC} \\ 
\multicolumn{1}{l|}{} & \multicolumn{1}{c|}{} & \multicolumn{1}{c|}{} & \multicolumn{1}{c|}{} & mAP & \multicolumn{1}{c|}{R1} & \multicolumn{1}{c|}{} & \multicolumn{1}{c|}{} & \multicolumn{1}{c|}{} & mAP & R1 \\ \hline
\multicolumn{1}{l|}{QAConv$_{50}$ \cite{liao2020interpretable}} & \multicolumn{1}{c|}{\multirow{4}{*}{D+C3+MT}} & \multicolumn{1}{c|}{\multirow{4}{*}{2,510}} & \multicolumn{1}{c|}{\multirow{4}{*}{56,508}} & 39.5 & \multicolumn{1}{c|}{68.6} & \multicolumn{1}{c|}{\multirow{4}{*}{M+C3+MT}} & \multicolumn{1}{c|}{\multirow{4}{*}{2,559}} & \multicolumn{1}{c|}{\multirow{4}{*}{52,922}} & 43.4 & 64.9 \\
\multicolumn{1}{l|}{M$^{3}$L \cite{zhao2021learning}} & \multicolumn{1}{c|}{} & \multicolumn{1}{c|}{} & \multicolumn{1}{c|}{} & \underline{51.1} & \multicolumn{1}{c|}{\underline{76.5}} & \multicolumn{1}{c|}{} & \multicolumn{1}{c|}{} & \multicolumn{1}{c|}{} & \underline{48.2} & \underline{67.1} \\
\multicolumn{1}{l|}{BoT \cite{luo2019strong}} & \multicolumn{1}{c|}{} & \multicolumn{1}{c|}{} & \multicolumn{1}{c|}{} &46.3  & \multicolumn{1}{c|}{75.1} & \multicolumn{1}{c|}{} & \multicolumn{1}{c|}{} & \multicolumn{1}{c|}{} &46.8  &65.3   \\
\multicolumn{1}{l|}{Our IDM \cite{dai2021idm}} & \multicolumn{1}{c|}{} & \multicolumn{1}{c|}{} & \multicolumn{1}{c|}{} &47.8  & \multicolumn{1}{c|}{75.8} & \multicolumn{1}{c|}{} & \multicolumn{1}{c|}{} & \multicolumn{1}{c|}{} &48.1  &66.3  \\
\multicolumn{1}{l|}{Our IDM++} & \multicolumn{1}{c|}{} & \multicolumn{1}{c|}{} & \multicolumn{1}{c|}{} &\textbf{52.6}  &\multicolumn{1}{c|}{\textbf{79.2}} & \multicolumn{1}{c|}{} & \multicolumn{1}{c|}{} & \multicolumn{1}{c|}{} &\textbf{52.7}  &\textbf{70.9}  \\ \hline
\multicolumn{1}{l|}{\multirow{2}{*}{Method}} & \multicolumn{1}{c|}{\multirow{2}{*}{Source}} & \multicolumn{1}{c|}{\multirow{2}{*}{IDs}} & \multicolumn{1}{c|}{\multirow{2}{*}{Images}} & \multicolumn{2}{c|}{Target: CUHK03} & \multicolumn{1}{c|}{\multirow{2}{*}{Source}} & \multicolumn{1}{c|}{\multirow{2}{*}{IDs}} & \multicolumn{1}{c|}{\multirow{2}{*}{Images}} & \multicolumn{2}{c}{Target: MSMT17} \\ 
\multicolumn{1}{l|}{} & \multicolumn{1}{c|}{} & \multicolumn{1}{c|}{} & \multicolumn{1}{c|}{} & mAP & \multicolumn{1}{c|}{R1} & \multicolumn{1}{c|}{} & \multicolumn{1}{c|}{} & \multicolumn{1}{c|}{} & mAP & R1 \\ \hline
\multicolumn{1}{l|}{QAConv$_{50}$ \cite{liao2020interpretable}} & \multicolumn{1}{c|}{\multirow{4}{*}{M+D+MT}} & \multicolumn{1}{c|}{\multirow{4}{*}{2,494}} & \multicolumn{1}{c|}{\multirow{4}{*}{62,079}} & 19.2 & \multicolumn{1}{c|}{22.9} & \multicolumn{1}{c|}{\multirow{4}{*}{M+D+C3}} & \multicolumn{1}{c|}{\multirow{4}{*}{2,220}} & \multicolumn{1}{c|}{\multirow{4}{*}{36,823}} & 10.0 & 29.9 \\
\multicolumn{1}{l|}{M$^{3}$L \cite{zhao2021learning}} & \multicolumn{1}{c|}{} & \multicolumn{1}{c|}{} & \multicolumn{1}{c|}{} & \underline{30.9} & \multicolumn{1}{c|}{\underline{31.9}} & \multicolumn{1}{c|}{} & \multicolumn{1}{c|}{} & \multicolumn{1}{c|}{} & 13.1 & 32.0 \\
\multicolumn{1}{l|}{BoT \cite{luo2019strong}} & \multicolumn{1}{c|}{} & \multicolumn{1}{c|}{} & \multicolumn{1}{c|}{} &26.0  & \multicolumn{1}{c|}{24.9} & \multicolumn{1}{c|}{} & \multicolumn{1}{c|}{} & \multicolumn{1}{c|}{} &12.6  &31.7   \\
\multicolumn{1}{l|}{Our IDM \cite{dai2021idm}} & \multicolumn{1}{c|}{} & \multicolumn{1}{c|}{} & \multicolumn{1}{c|}{} &27.8  & \multicolumn{1}{c|}{27.7} & \multicolumn{1}{c|}{} & \multicolumn{1}{c|}{} & \multicolumn{1}{c|}{} &\underline{14.8}  &\underline{36.5}  \\
\multicolumn{1}{l|}{Our IDM++} & \multicolumn{1}{c|}{} & \multicolumn{1}{c|}{} & \multicolumn{1}{c|}{} &\textbf{32.7}  &\multicolumn{1}{c|}{\textbf{33.6}} & \multicolumn{1}{c|}{} & \multicolumn{1}{c|}{} & \multicolumn{1}{c|}{} &\textbf{19.5}  &\textbf{44.3}  \\ \hline \hline
\multicolumn{11}{c}{DG-full} \\ \hline
\multicolumn{1}{l|}{\multirow{2}{*}{Method}} & \multicolumn{1}{c|}{\multirow{2}{*}{Source}} & \multicolumn{1}{c|}{\multirow{2}{*}{IDs}} & \multicolumn{1}{c|}{\multirow{2}{*}{Images}} & \multicolumn{2}{c|}{Target: Market-1501} & \multicolumn{1}{c|}{\multirow{2}{*}{Source}} & \multicolumn{1}{c|}{\multirow{2}{*}{IDs}} & \multicolumn{1}{c|}{\multirow{2}{*}{Images}} & \multicolumn{2}{c}{Target: DukeMTMC} \\ 
\multicolumn{1}{l|}{} & \multicolumn{1}{c|}{} & \multicolumn{1}{c|}{} & \multicolumn{1}{c|}{} & mAP & \multicolumn{1}{c|}{R1} & \multicolumn{1}{c|}{} & \multicolumn{1}{c|}{} & \multicolumn{1}{c|}{} & mAP & R1 \\ \hline
\multicolumn{1}{l|}{RaMoE \cite{dai2021generalizable}} & \multicolumn{1}{c|}{\multirow{3}{*}{D+C3+MT}} & \multicolumn{1}{c|}{\multirow{3}{*}{7,380}} & \multicolumn{1}{c|}{\multirow{3}{*}{176,949}} & \underline{56.5} & \multicolumn{1}{c|}{\underline{82.0}} & \multicolumn{1}{c|}{\multirow{3}{*}{M+C3+MT}} & \multicolumn{1}{c|}{\multirow{3}{*}{7,069}} & \multicolumn{1}{c|}{\multirow{3}{*}{169,957}} & \underline{56.9} & \underline{73.6} \\
\multicolumn{1}{l|}{BoT \cite{luo2019strong}} & \multicolumn{1}{c|}{} & \multicolumn{1}{c|}{} & \multicolumn{1}{c|}{} &50.2  & \multicolumn{1}{c|}{77.0} & \multicolumn{1}{c|}{} & \multicolumn{1}{c|}{} & \multicolumn{1}{c|}{} &52.6  &68.4   \\
\multicolumn{1}{l|}{Our IDM \cite{dai2021idm}} & \multicolumn{1}{c|}{} & \multicolumn{1}{c|}{} & \multicolumn{1}{c|}{} &52.2  & \multicolumn{1}{c|}{77.3} & \multicolumn{1}{c|}{} & \multicolumn{1}{c|}{} & \multicolumn{1}{c|}{} &53.1  &70.0  \\
\multicolumn{1}{l|}{Our IDM++} & \multicolumn{1}{c|}{} & \multicolumn{1}{c|}{} & \multicolumn{1}{c|}{} &\textbf{58.8}  & \multicolumn{1}{c|}{\textbf{82.2}} & \multicolumn{1}{c|}{} & \multicolumn{1}{c|}{} & \multicolumn{1}{c|}{} &\textbf{58.3}  &\textbf{74.5}  \\ \hline
\multicolumn{1}{l|}{\multirow{2}{*}{Method}} & \multicolumn{1}{c|}{\multirow{2}{*}{Source}} & \multicolumn{1}{c|}{\multirow{2}{*}{IDs}} & \multicolumn{1}{c|}{\multirow{2}{*}{Images}} & \multicolumn{2}{c|}{Target: CUHK03} & \multicolumn{1}{c|}{\multirow{2}{*}{Source}} & \multicolumn{1}{c|}{\multirow{2}{*}{IDs}} & \multicolumn{1}{c|}{\multirow{2}{*}{Images}} & \multicolumn{2}{c}{Target: MSMT17} \\ 
\multicolumn{1}{l|}{} & \multicolumn{1}{c|}{} & \multicolumn{1}{c|}{} & \multicolumn{1}{c|}{} & mAP & \multicolumn{1}{c|}{R1} & \multicolumn{1}{c|}{} & \multicolumn{1}{c|}{} & \multicolumn{1}{c|}{} & mAP & R1 \\ \hline
\multicolumn{1}{l|}{RaMoE \cite{dai2021generalizable}} & \multicolumn{1}{c|}{\multirow{3}{*}{M+D+MT}} & \multicolumn{1}{c|}{\multirow{3}{*}{7,414}} & \multicolumn{1}{c|}{\multirow{3}{*}{192,271}} & 35.5 & \multicolumn{1}{c|}{\underline{36.6}} & \multicolumn{1}{c|}{\multirow{3}{*}{M+D+C3}} & \multicolumn{1}{c|}{\multirow{3}{*}{4,780}} & \multicolumn{1}{c|}{\multirow{3}{*}{79,927}} & 13.5 & 34.1 \\
\multicolumn{1}{l|}{BoT \cite{luo2019strong}} & \multicolumn{1}{c|}{} & \multicolumn{1}{c|}{} & \multicolumn{1}{c|}{} &30.5  & \multicolumn{1}{c|}{29.9} & \multicolumn{1}{c|}{} & \multicolumn{1}{c|}{} & \multicolumn{1}{c|}{} &12.0  &29.9   \\
\multicolumn{1}{l|}{Our IDM \cite{dai2021idm}} & \multicolumn{1}{c|}{} & \multicolumn{1}{c|}{} & \multicolumn{1}{c|}{} &\underline{35.7}  & \multicolumn{1}{c|}{36.0} & \multicolumn{1}{c|}{} & \multicolumn{1}{c|}{} & \multicolumn{1}{c|}{} &\underline{15.3}  &\underline{36.7}  \\
\multicolumn{1}{l|}{Our IDM++} & \multicolumn{1}{c|}{} & \multicolumn{1}{c|}{} & \multicolumn{1}{c|}{} &\textbf{39.9}  & \multicolumn{1}{c|}{\textbf{40.9}} & \multicolumn{1}{c|}{} & \multicolumn{1}{c|}{} & \multicolumn{1}{c|}{} &\textbf{20.1}  &\textbf{44.3}  \\ \hline
\end{tabular}
\end{center}
\end{table*}

\textbf{Comparisons on synthetic $\to$ real UDA re-ID tasks.} Compared with the real $\to$ real UDA re-ID tasks, the synthetic $\to$ real UDA tasks are more challenging because the domain gap between the synthetic and real images are often larger than that between the real and real images. As shown in Table \ref{tab:SOTA-synthetic2real}, our method can outperform the SOTA methods by a large margin where mAP of our method is higher than CCL~\cite{isobe2021towards} by 4.7\% and 8.8\% when transferring from PersonX to Market-1501 and MSMT17 respectively. 
All these significant performance gains in Table~\ref{tab:SOTA-synthetic2real} have shown the superiority of our method that can better improve the performance on UDA re-ID by extending IDM to IDM++.

\textbf{Comparisons on DG re-ID tasks.}
In DG re-ID tasks, we utilize BoT \cite{luo2019strong} as the baseline method and it is competitive to the existing methods \cite{liao2020interpretable,zhao2021heterogeneous,dai2021generalizable}. Based on this baseline method, we implement our method IDM++ and our previous work IDM. As shown in Table \ref{tab:SOTA-DG-reid}, we compare our method with the SOTAs under two common protocols: \textit{i.e.,} DG-partial~\cite{zhao2021heterogeneous} and DG-full \cite{dai2021generalizable}. Under the ``DG-partial'' protocol, mAP of our IDM++ outperforms the SOTA method M$^{3}$L \cite{zhao2021heterogeneous} by 1.5\%, 4.5\%, 1.8\%, and 6.4\% when testing on Market-1501, DukeMTMC, CUHK03, and MSMT17 respectively. Under the ``DG-full'' protocol, our method outperforms the SOTA method RaMoE \cite{dai2021generalizable} significantly as well. Compared with our previous work (IDM \cite{dai2021idm}), the performance gain of our IDM++ is significant. Taking the protocol of ``DG-ful'' as an example, mAP of IDM++ outperforms IDM by 6.6\%, 5.2\%, 4.2\%, and 4.8\% when testing on Market-1501, DukeMTMC, CUHK03, and MSMT17 respectively. 
In DG re-ID, the performance gain mainly comes from our MGM.
The IDM method is superior to the baseline BoT because mixing source and target domains to generate intermediate domains can be seen as only considering inter-domain knowledge. By reinforcing IDM into IDM++ by adding MGM, features are learned to be more identity-discriminative and more robust to cross-domain variations.

\subsection{Investigation on Key Designs of IDM++}
\label{sec:investigation_on_mechanism}
\begin{figure}[tp]
		\centering
		\scriptsize
		\setlength\tabcolsep{1mm}
		\renewcommand\arraystretch{0.1}
		\begin{tabular}{cc}
			\includegraphics[width=0.46\linewidth]{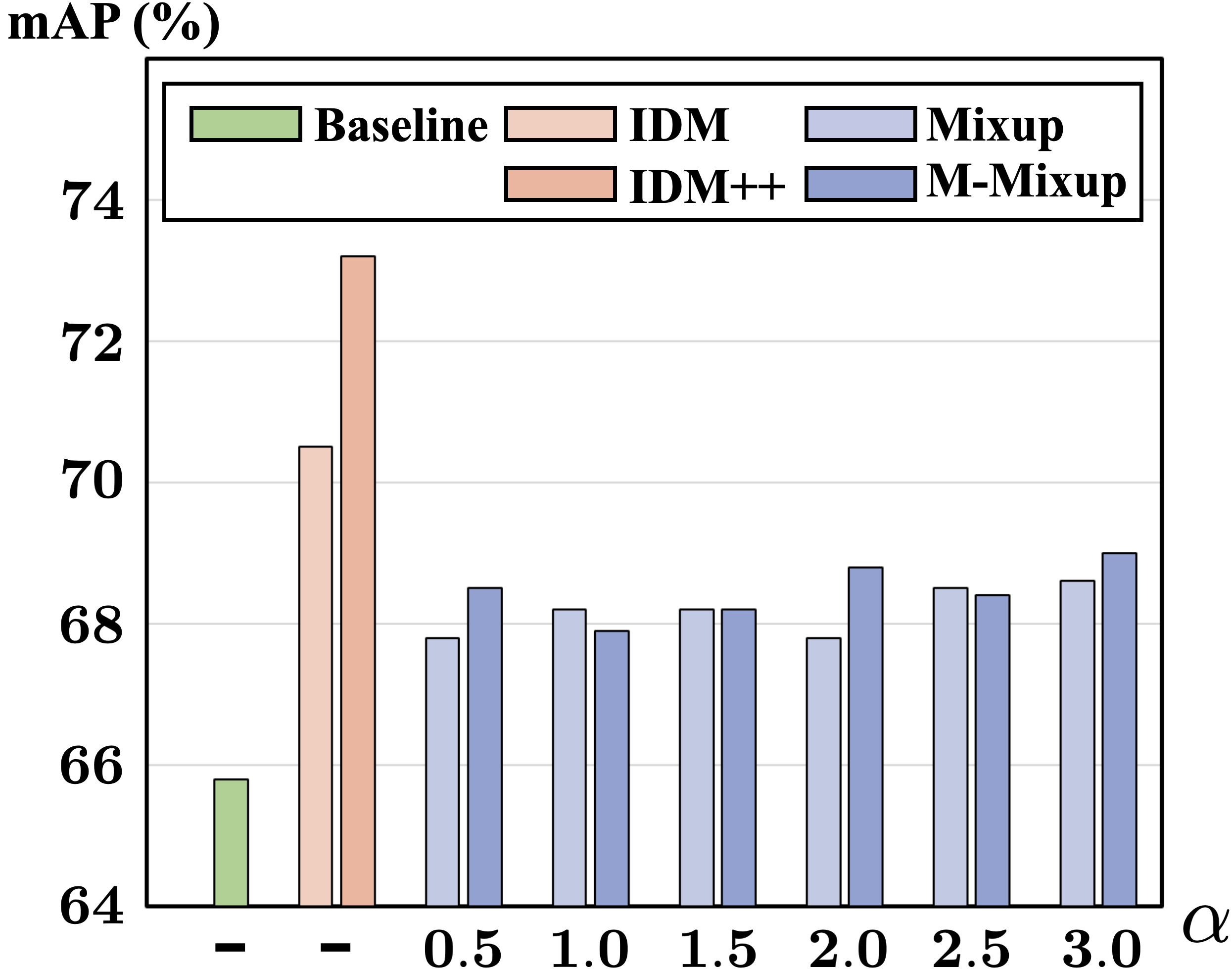} &
			\includegraphics[width=0.46\linewidth]{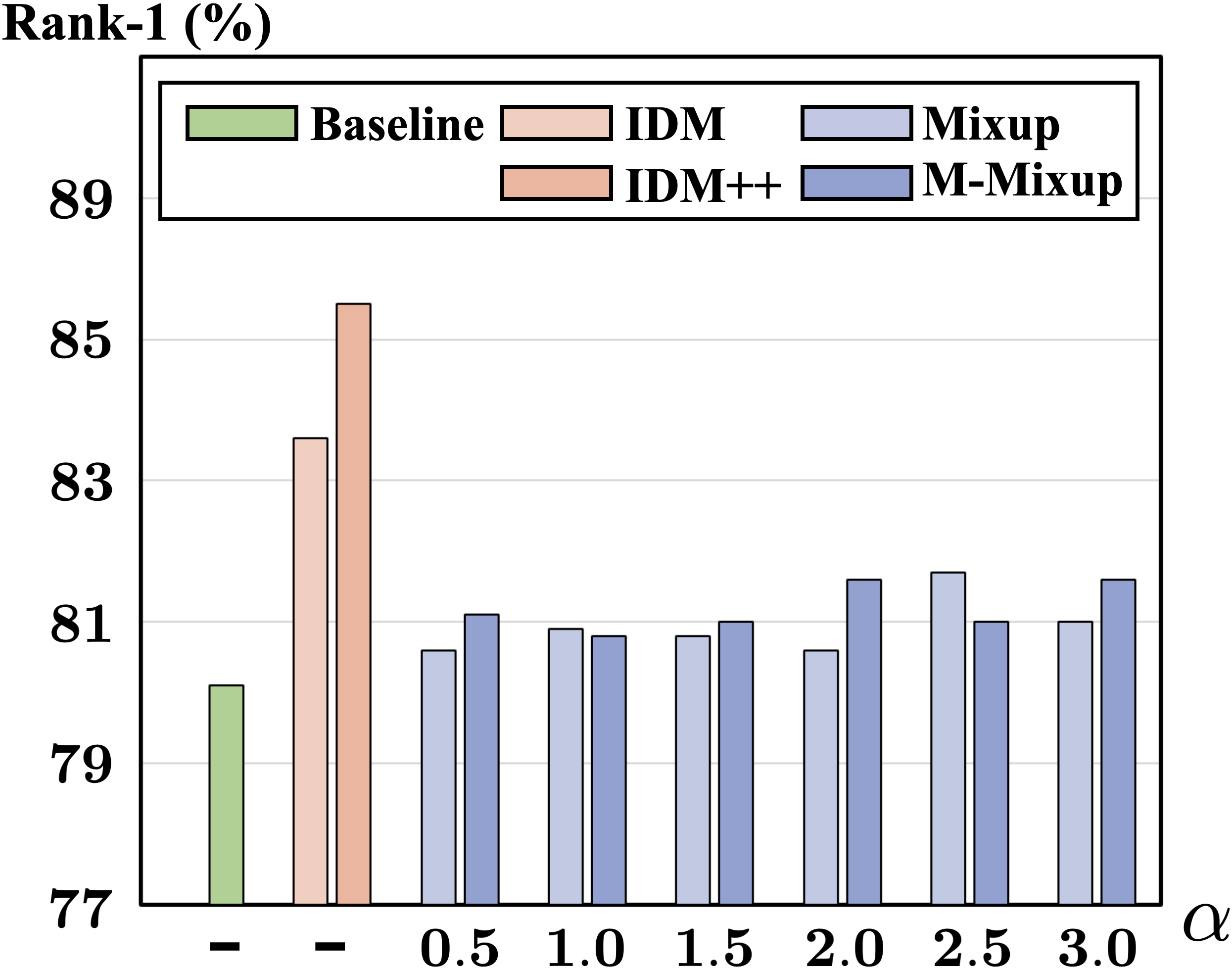} \\
			~\\
			(a) mAP & (b) Rank-1
		\end{tabular}
		\caption{Comparison with traditional mixup methods (Mixup \cite{zhang2017mixup}, M-Mixup \cite{verma2019manifold}) on Market$\to$Duke. The interpolation ratio in these mixup methods is randomly sampled from a beta distribution ${\rm Beta}(\alpha,\alpha)$.}
		\label{fig:vs_mixup}
\end{figure}

\textbf{The mixup operation in IDM.} We compare our method with the traditional Mixup \cite{zhang2017mixup} and Manifold Mixup (M-Mixup) \cite{verma2019manifold} methods in Fig.~\ref{fig:vs_mixup}. We use Mixup to randomly mix the source and target domains at the image-level, and use M-Mixup to randomly mix the two domains at the feature-level. The interpolation ratio in Mixup and M-Mixup is randomly sampled from a beta distribution ${\rm Beta}(\alpha,\alpha)$. Specifically, Mixup and M-Mixup are only the image/feature augmentation technology and we use them to randomly mix the source and target domains. Unlike them, our method can adaptively generate the mix ratios to mix up the source and target domains representations to generate intermediate domain representations. Thus, our method is superior to these mixup methods in UDA re-ID.

\begin{table}[tp]
\centering
\caption{Comparisons on different mirrors generation mechanisms. The mirrors-I means using MGM to map source and target samples into intermediate domains. The mirrors-ST means mapping source and target samples into each other's domain.}
\label{tab:alignment_mechanism}
\begin{tabular}{l|cc}
\hline
\multirow{2}{*}{Mirrors Generation Mechanism} & \multicolumn{2}{c}{Market $\to$ Duke} \\ \cline{2-3} 
 & mAP & R1 \\ \hline
IDM+MGM w/ mirrors-I &73.2  &85.5  \\
IDM+MGM w/ mirrors-ST &72.5  &83.8  \\
MGM w/ mirrors-ST &68.0  &81.7  \\ 
\hline
\end{tabular}
\end{table}

\begin{figure}[tp]
		\centering
		\scriptsize
		\setlength\tabcolsep{1mm}
		\renewcommand\arraystretch{0.1}
		\begin{tabular}{cc}
			\includegraphics[width=0.46\linewidth]{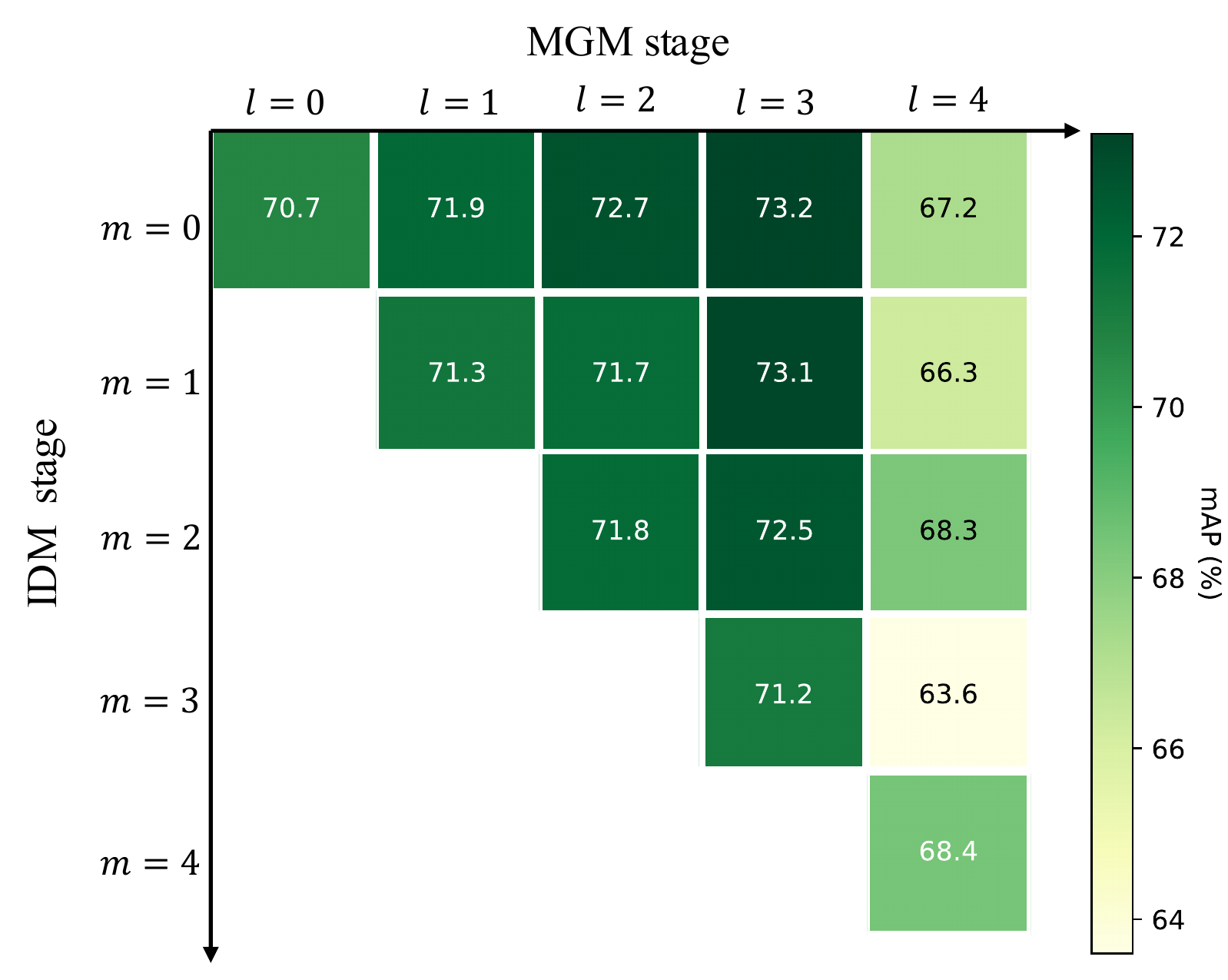} &
			\includegraphics[width=0.46\linewidth]{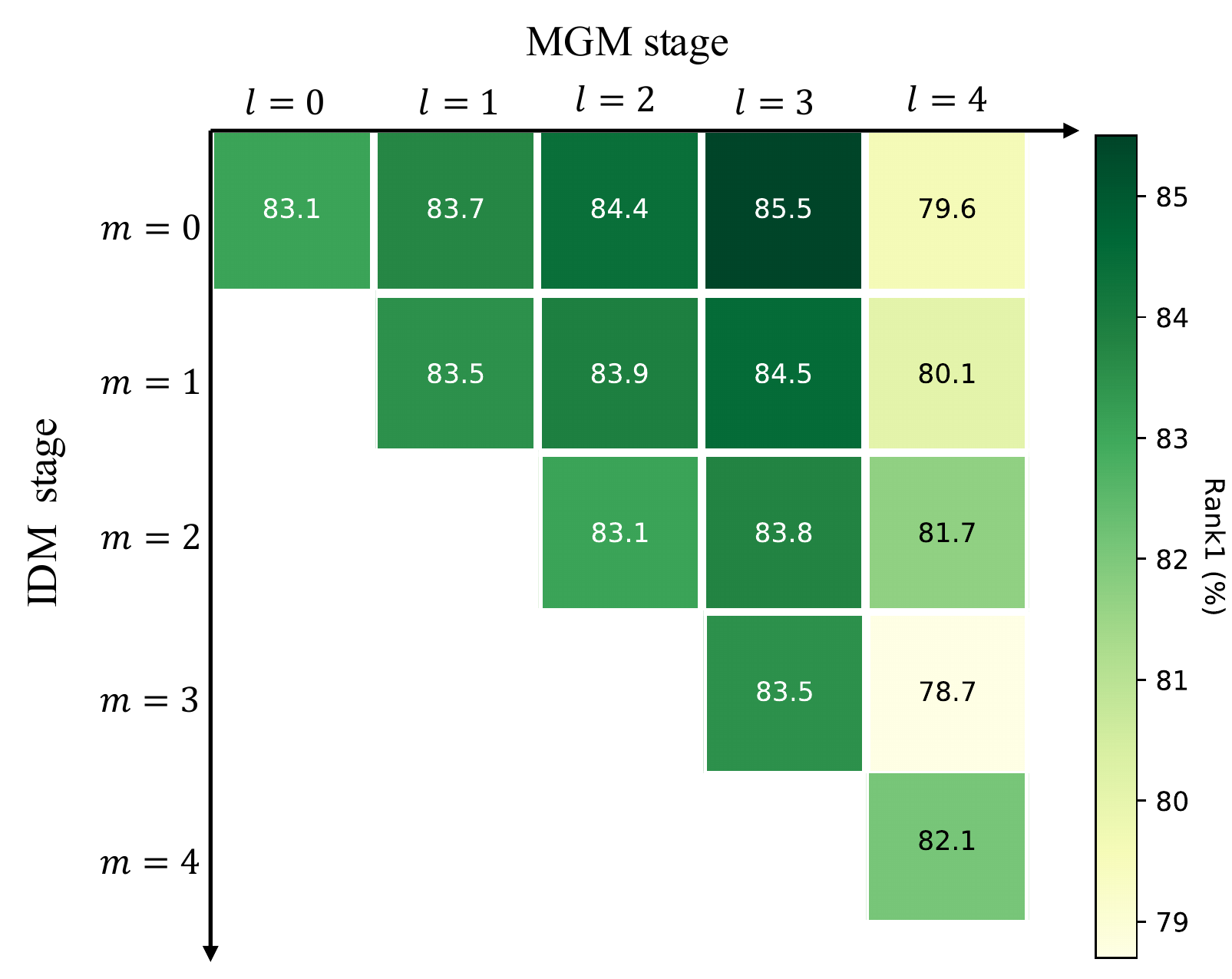} \\
			~\\
			(a) mAP & (b) Rank-1
		\end{tabular}
		\caption{Evaluation on which stage (stage-$m$) to plug our IDM and which stage (stage-$l$) to plug our MGM. We provide the mAP and Rank-1 on Market$\to$Duke.}
		\label{fig:ablation_stage}
\end{figure}

\textbf{The mirrors generation mechanism in MGM.} We compare with different mirrors generation mechanisms in Table~\ref{tab:alignment_mechanism}. In our IDM++, we use the mechanism ``IDM+MGM w/ mirrors-I'' as shown in Section~\ref{sec:domain_mapping}. If we use the MGM to map source and target features into each other's domain (``mirrors-ST''), the inputs of the MGM (as shown in Fig.~\ref{fig:pipeline}) are only the source and target hidden representations (\textit{i.e.,} $G^{s}$ and $G^{t}$). Thus, we can obtain the source mirrors $G^{s \rightarrow t}$ and the target mirrors $G^{t \rightarrow s}$ that are stylized by the target and source domains respectively. In Table~\ref{tab:alignment_mechanism}, the mechanism ``MGM w/ mirrors-ST'' means directly aligning between the source and target domains, which is inferior to aligning by intermediate domains. The mechanism ``IDM+MGM w/ mirrors-ST'' means utilizing the IDM to generate intermediate domains but mapping source / target identities into each other other's domain to obtain mirrors. From Table~\ref{tab:alignment_mechanism}, we can see the significance of our proposed mirrors generation mechanism.

\textbf{Plugging IDM and MGM at which stage.}
Our IDM and MGM are two plug-and-play modules which can be seamlessly plugged into the backbone network.
In our experiments, we use ResNet-50 as the backbone, which has five stages: stage-0 is comprised of the first Conv, BN and Max Pooling layer and stage-1/2/3/4 correspond to the other four convolutional blocks. We plug our IDM at the $m$-th stage and plug our MGM at the $l$-th stage to study how different combinations of both kinds of stages will affect the performance of our method. 
The stage-$l$ of MGM should not be smaller than the stage-$m$ of IDM because IDM is a prerequisite for MGM (\textit{i.e.,} MGM needs to map source / target identities into intermediate domains generated by IDM).
As shown in Fig.~\ref{fig:ablation_stage}, we provide the visualization of the target domain's mAP/Rank-1 on different combinations of the stage-$m$ and the stage-$l$ when transferring from Market-1501 to DukeMTMC. The Fig.~\ref{fig:ablation_stage} shows that the performance is gradually declining with the deepening of the network (\textit{i.e.,} $m$ ranges from 0 to 4). This phenomenon shows that the domain gap becomes larger and the transferable ability becomes weaker at higher/deeper layers of the network, which satisfies the theory of the domain adaptation~\cite{long2018transferable}. When the stage-$l$ of MGM ranges from 0 to 3, the performance becomes better, which shows that our proposed MGM can complement with the IDM in deeper layers that have weak transferable ability. We observe that when $l$ is 4, the model will be degenerated because features after stage-4 will be directly conducted with global average pooling to learn with the triplet loss. Compared  with shallow layers' features, the little manipulation on stage-4 features will make a significant effect on the deep embedding learning. Based on the above analysis, we plug the IDM at the stage-0 and plug the MGM at the stage-3 in all our experiments if not specified.

\begin{figure}[tp]
		\centering
		\scriptsize
		\setlength\tabcolsep{1mm}
		\renewcommand\arraystretch{0.1}
		\begin{tabular}{cc}
			\includegraphics[width=0.46\linewidth]{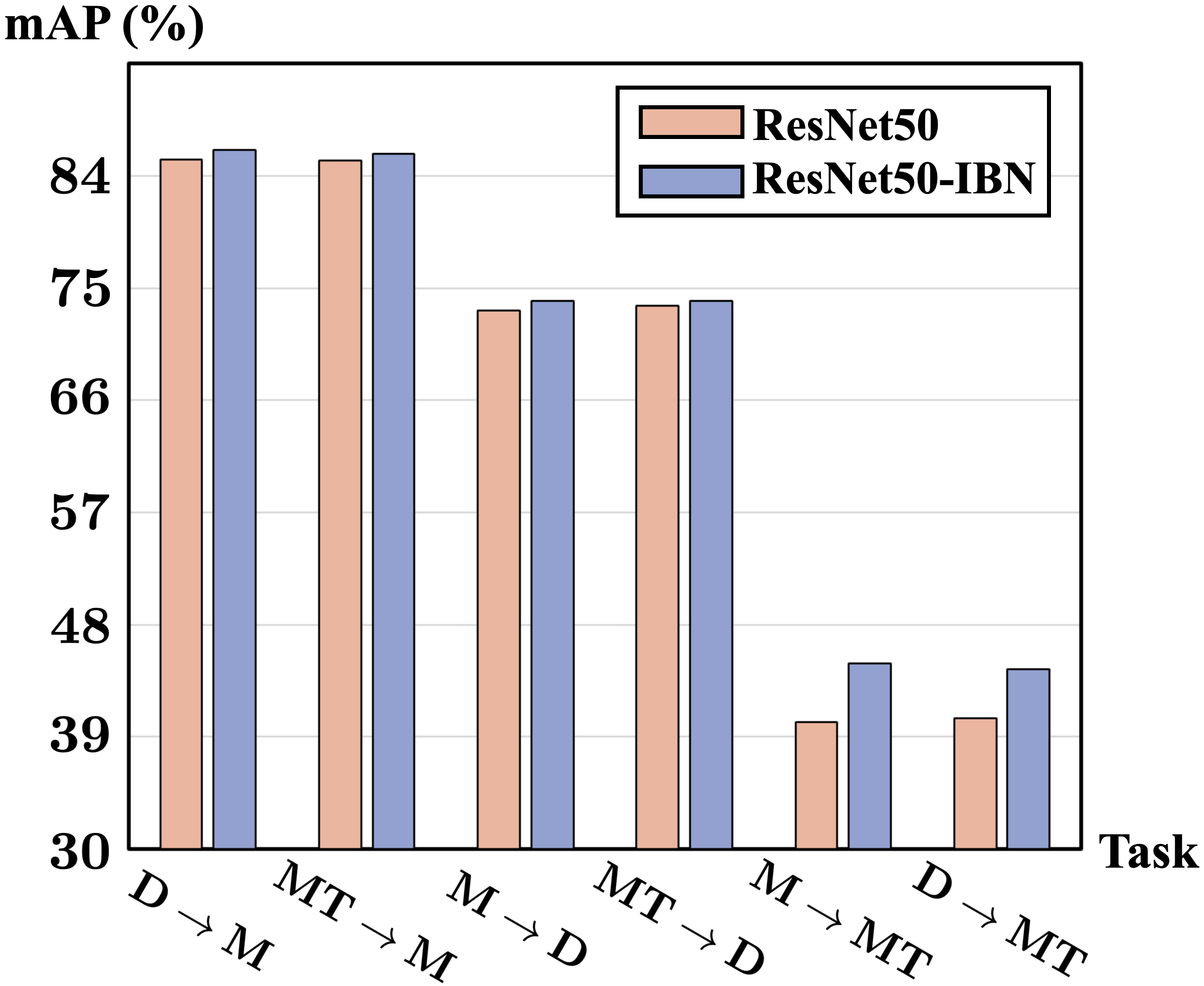} &
			\includegraphics[width=0.46\linewidth]{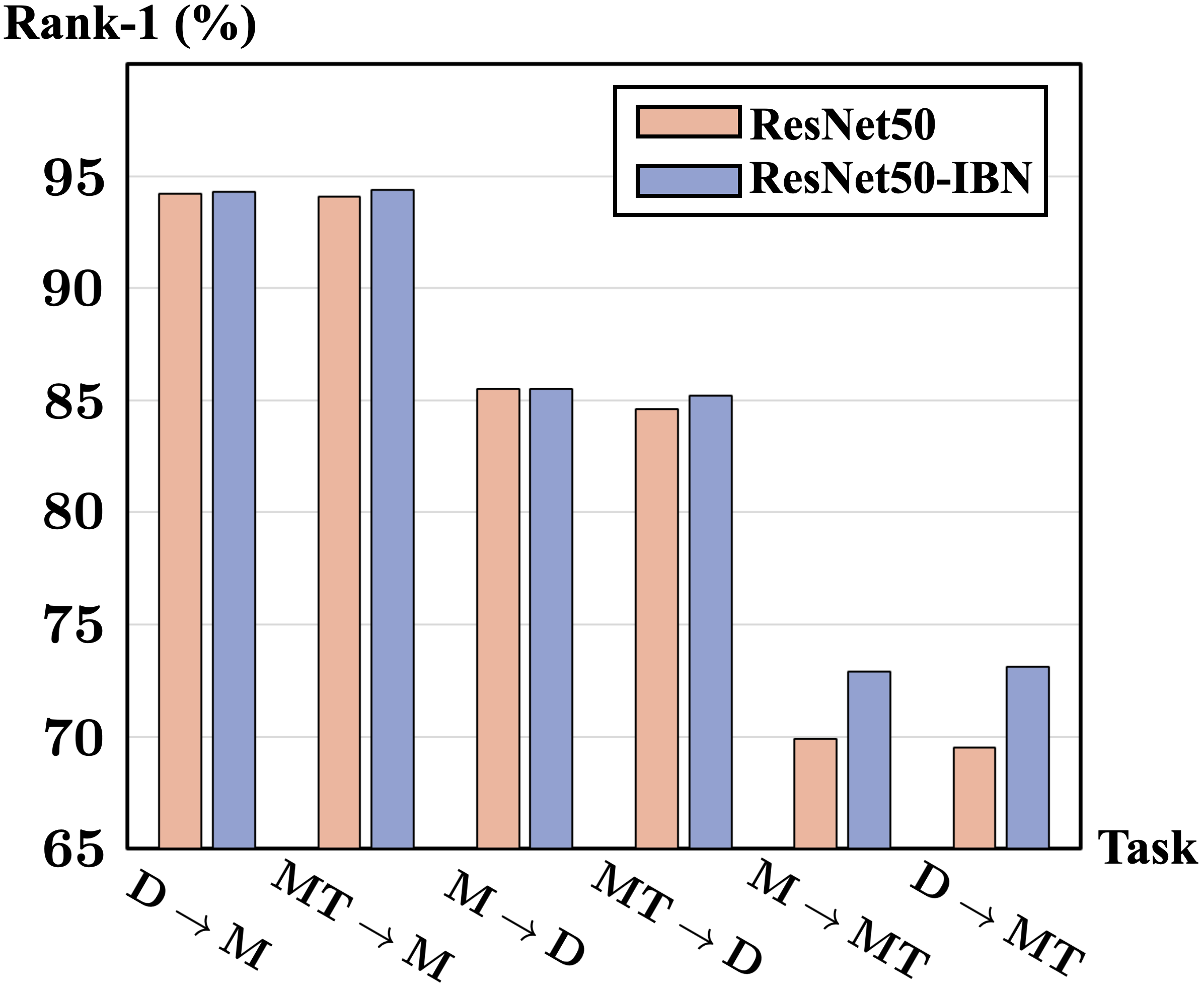} \\
			~\\
			(a) mAP & (b) Rank-1
		\end{tabular}
		\caption{Comparisons between different backbones: ResNet50 and ResNet50-IBN. We denote Market-1501, DukeMTMC-reID, and MSMT17 as M, D, and MT respectively.}
		\label{fig:ibn_map_r1}
\end{figure}

\begin{figure*}[htp]
	\centering
	\footnotesize
	\setlength\tabcolsep{1mm}
	\renewcommand\arraystretch{0.1}
	\begin{tabular}{ccc}
		\includegraphics[width=0.3\linewidth]{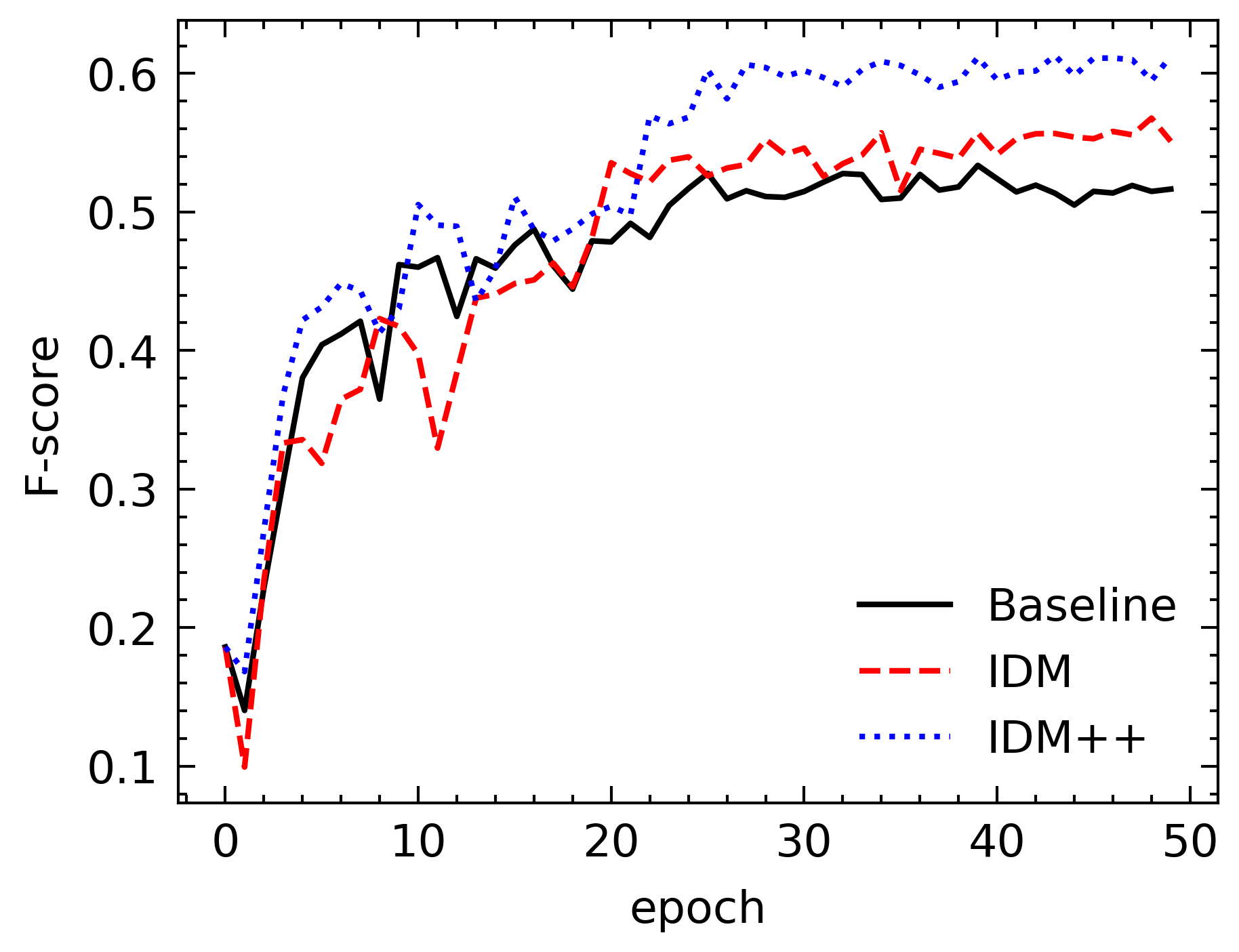} &
		\includegraphics[width=0.31\linewidth]{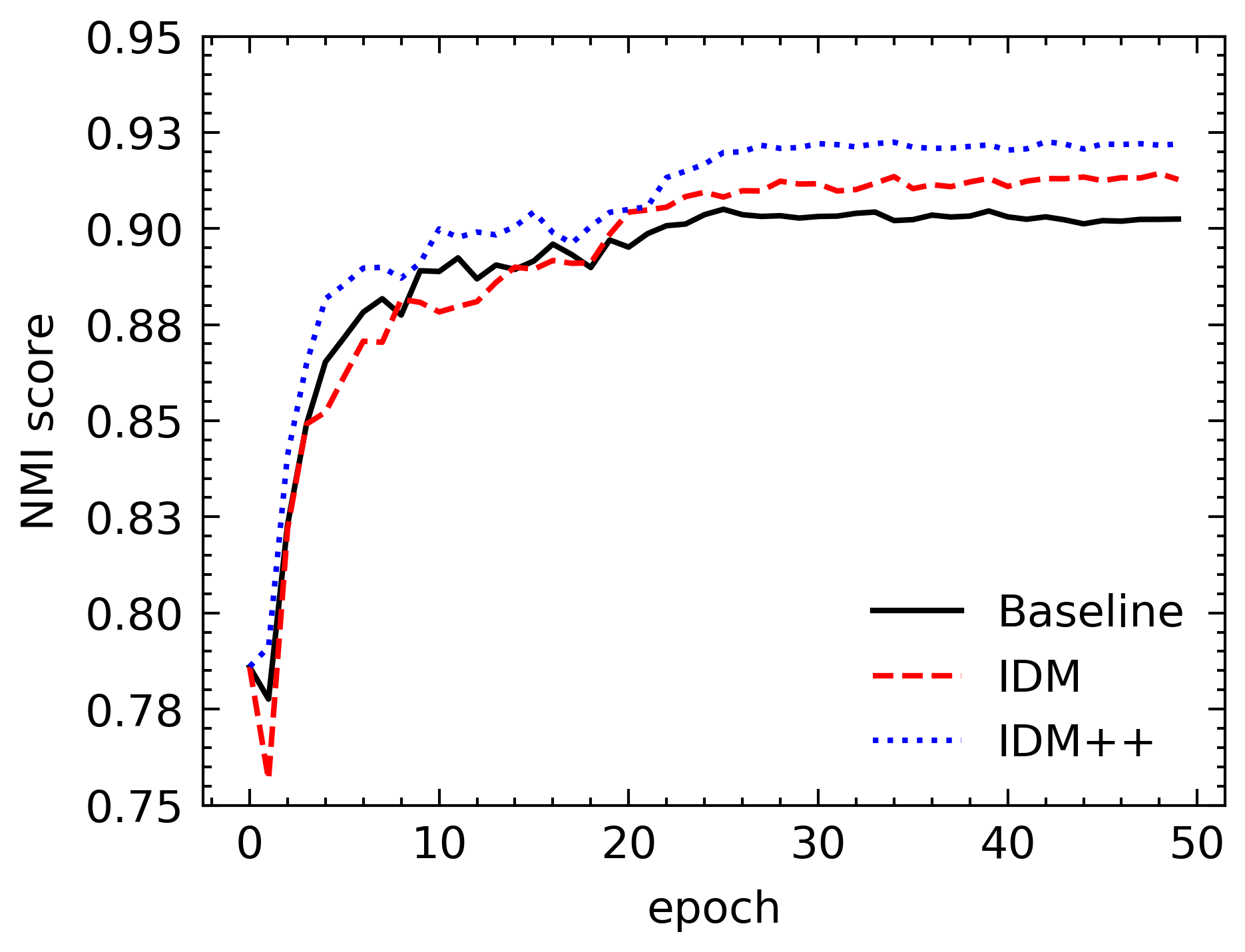} &
		\includegraphics[width=0.3\linewidth]{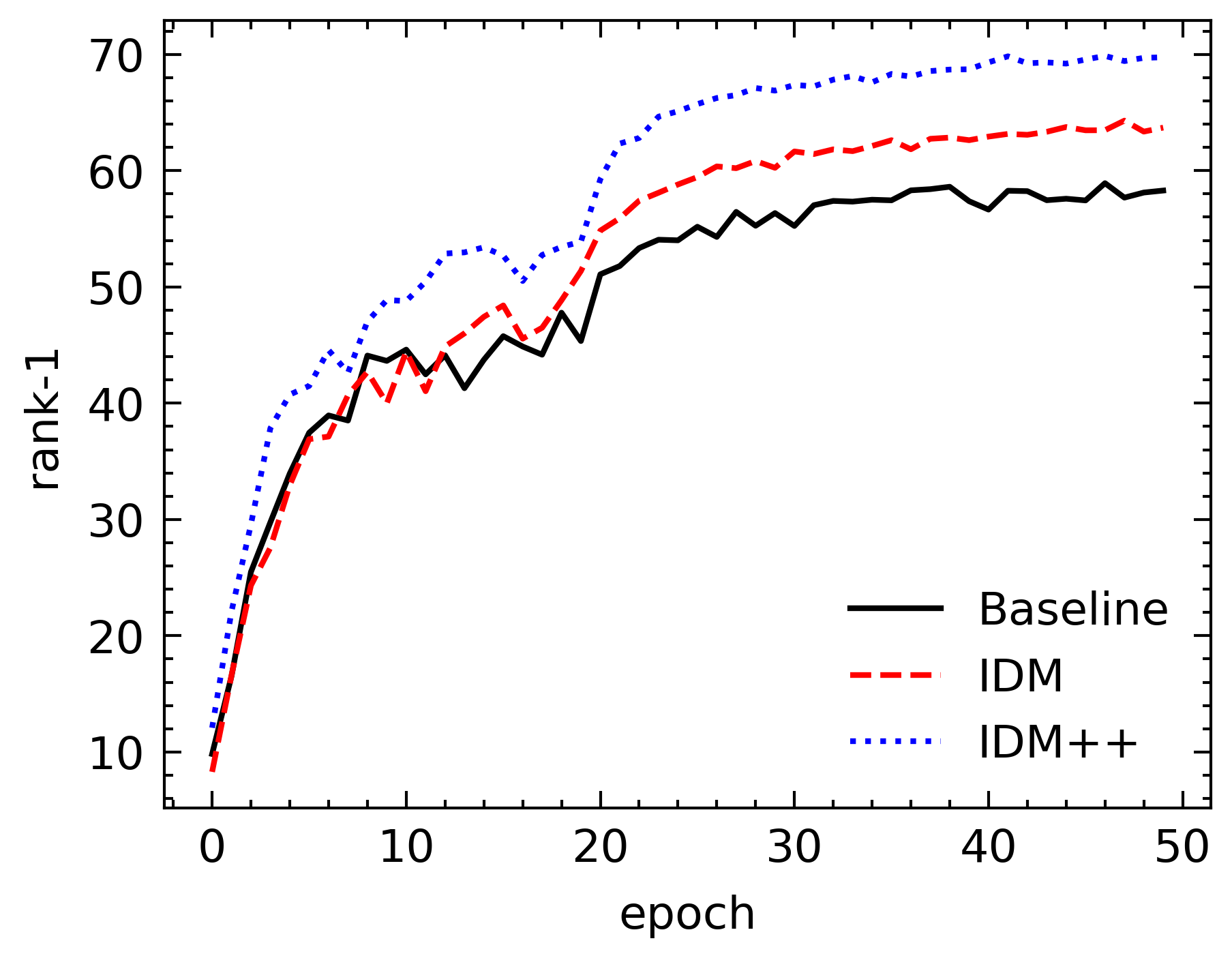} \\
		~\\
		(a) F-score & (b) NMI score & (c) Rank-1
	\end{tabular}
	\caption{Analysis on the F-score, NMI score, and Rank-1 score when training on Market-1501 and testing on MSMT17.}
	\label{fig:pseudo_labels}
\end{figure*} 

\begin{figure*}[htp]
	\centering
	\footnotesize
	\setlength\tabcolsep{1mm}
	\renewcommand\arraystretch{0.1}
	\begin{tabular}{cccc}
		\includegraphics[width=0.23\linewidth]{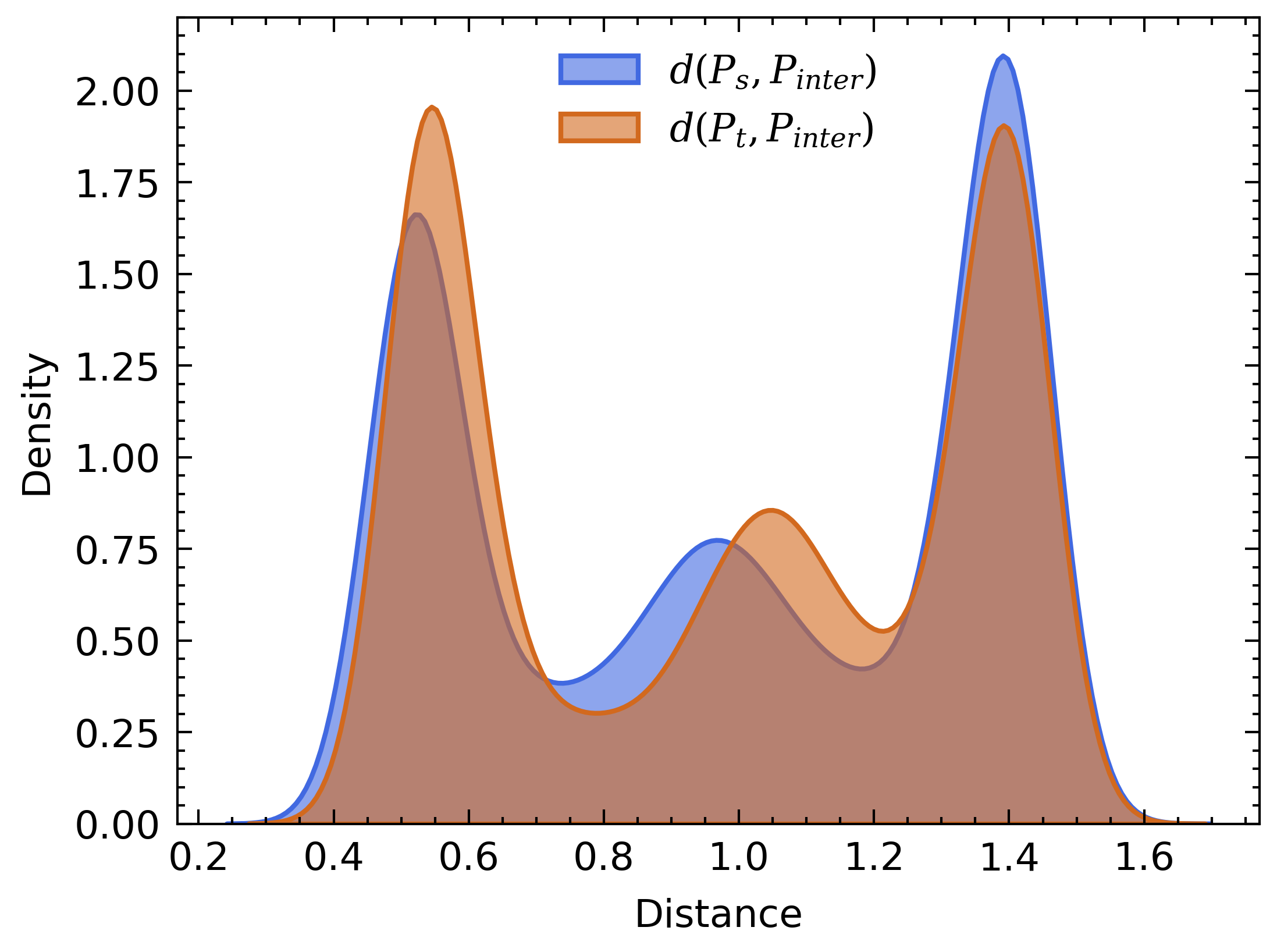} &
		\includegraphics[width=0.23\linewidth]{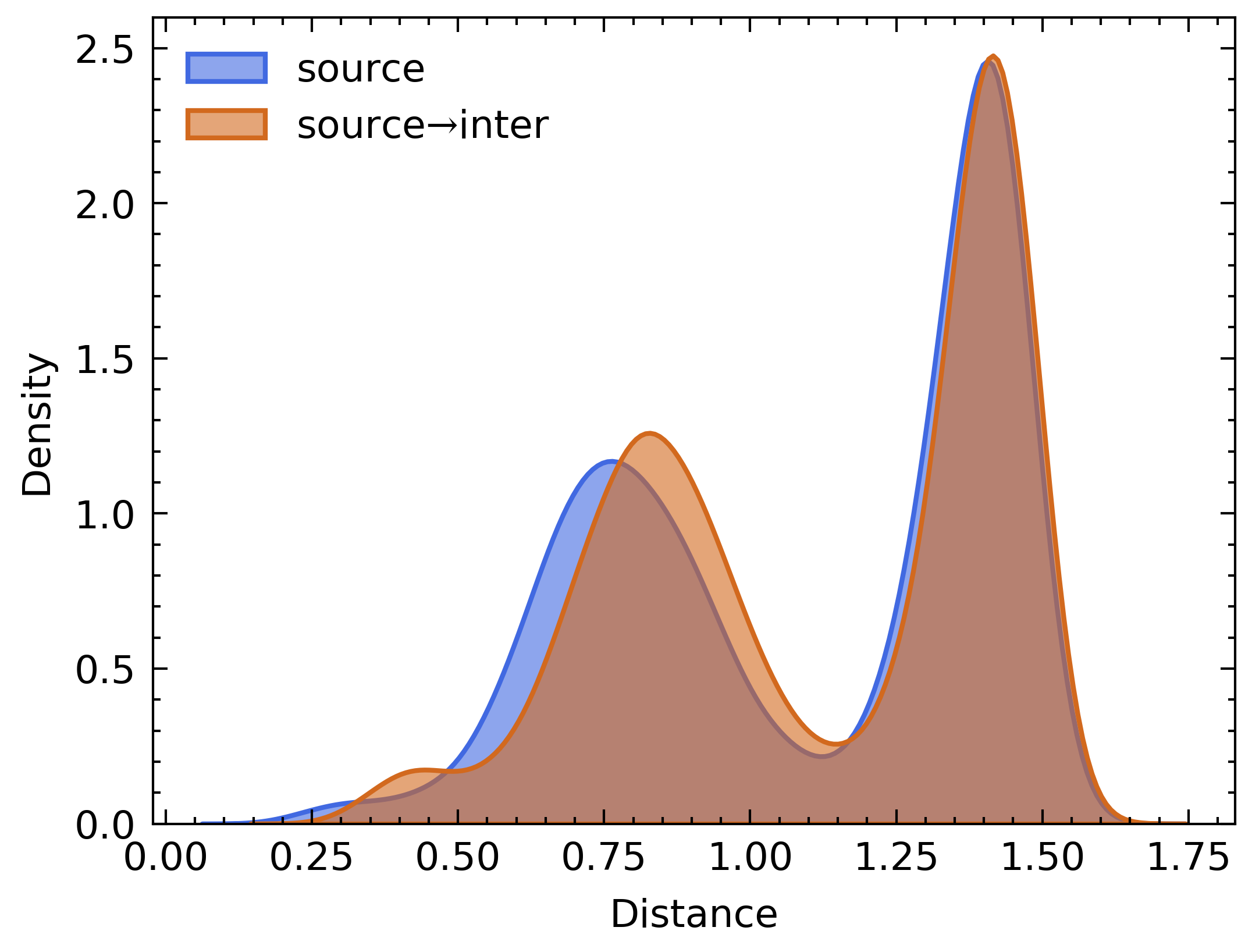} &
		\includegraphics[width=0.23\linewidth]{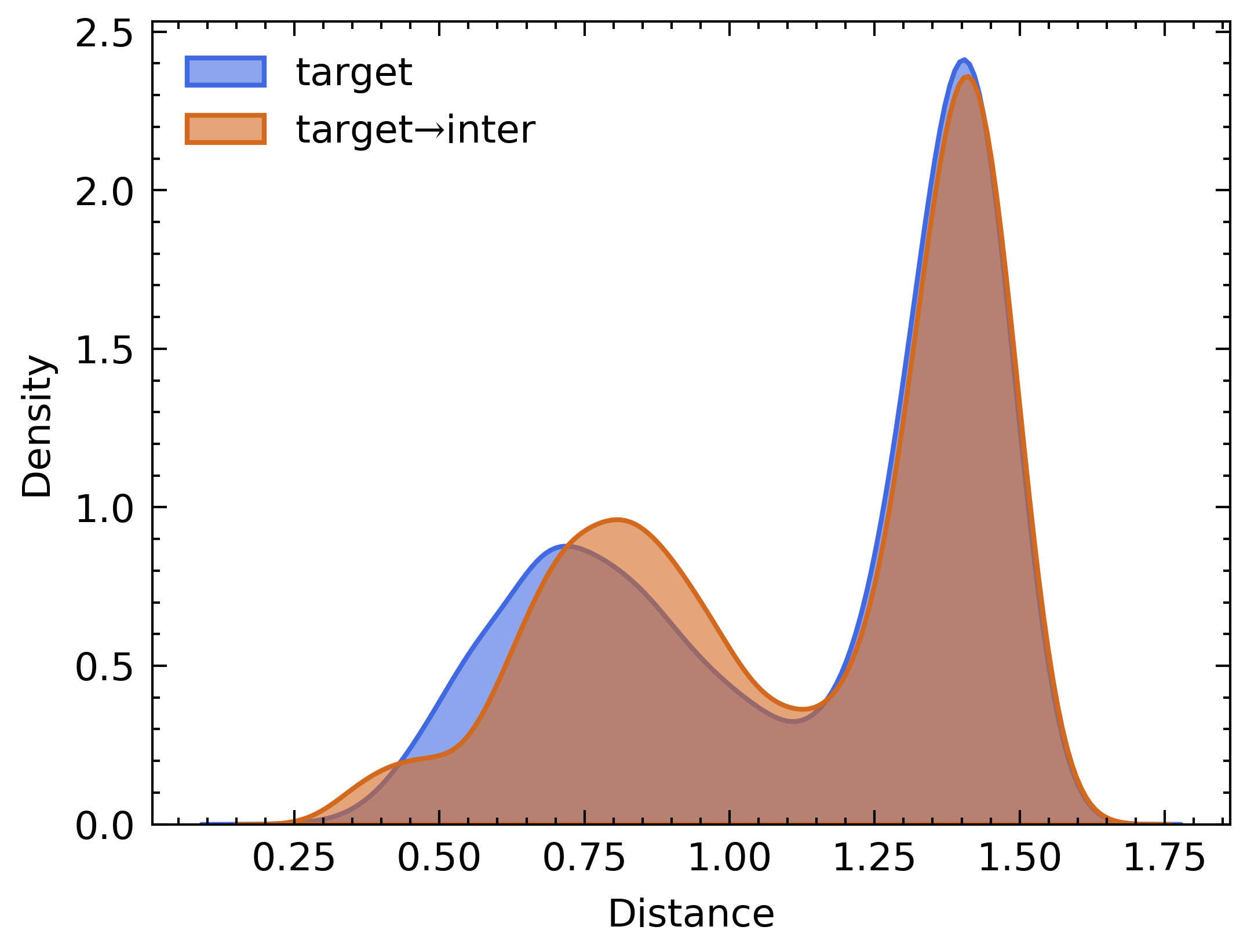} &
		\includegraphics[width=0.23\linewidth]{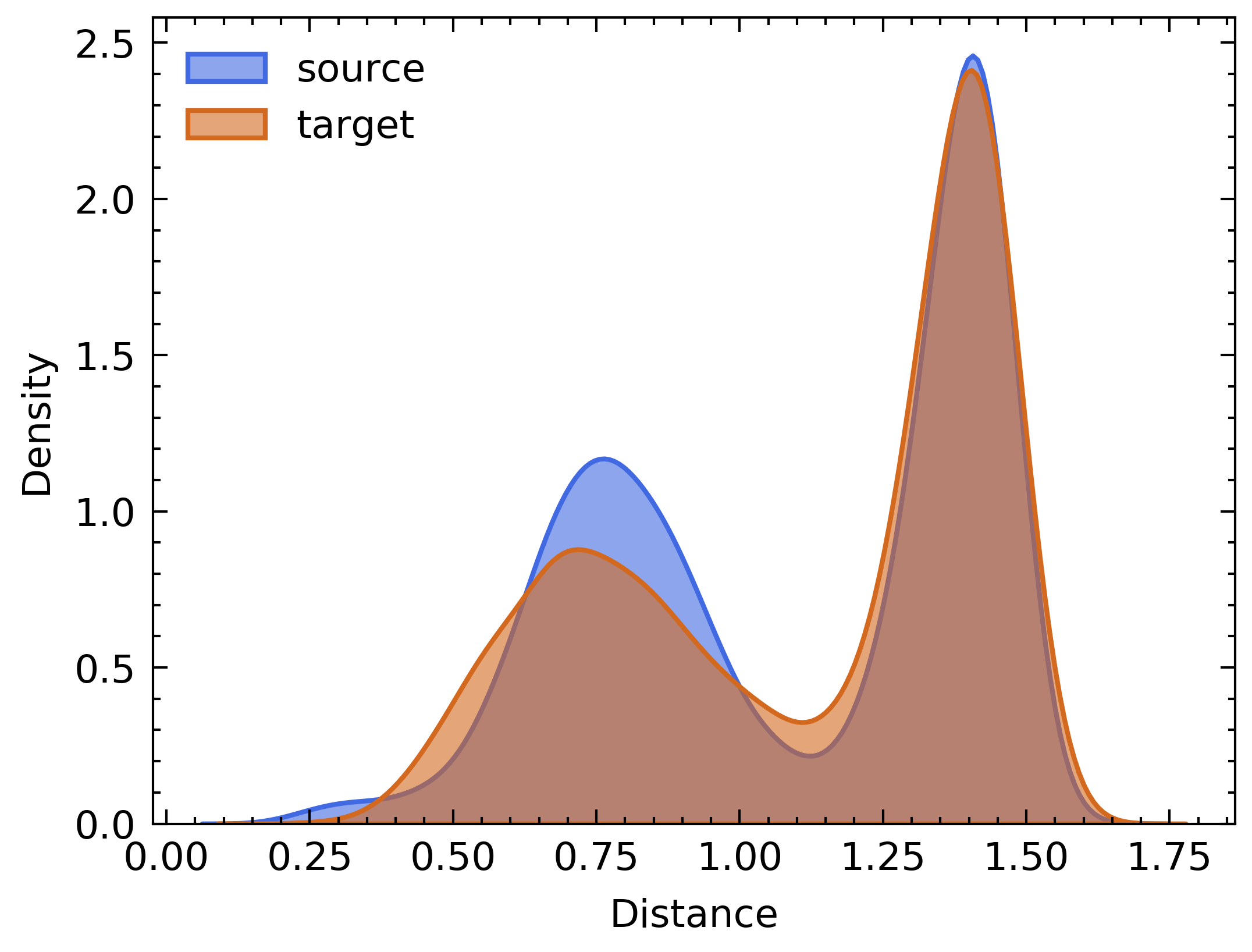} \\
		~\\
		(a) $d(P_{s}, P_{\rm inter}) \ \& \ d(P_{t}, P_{\rm inter})$ & (b) $P_{s} \ \& \ P_{s \to \rm inter}$ & (c) $P_{t} \ \& \ P_{t \to \rm inter}$ & (d) $P_{s} \ \& \ P_{t}$
	\end{tabular}
	\caption{
		Distribution of samples' pair-wise distance (Euclidean distance of the deep embedding) on the target dataset DukeMTMC-reID when taking Market-1501 as the source dataset. (a) Distance between source and intermediate domains $d(P_{s}, P_{\rm inter})$, and distance between target and intermediate domains $d(P_{t}, P_{\rm inter})$. (b) Distribution of source samples $P_{s}$ and their mirrors $P_{s \to \rm inter}$ in intermediate domains. (c) Distribution of target samples $P_{t}$ and their mirrors $P_{t \to \rm inter}$ in intermediate domains. (d) Distribution of source and target samples.}
	\label{fig:kde_all}
\end{figure*}

\begin{figure*}[htp]
	\centering
	\footnotesize
	\setlength\tabcolsep{1mm}
	\renewcommand\arraystretch{0.1}
	\begin{tabular}{cccc}
		\includegraphics[width=0.23\linewidth]{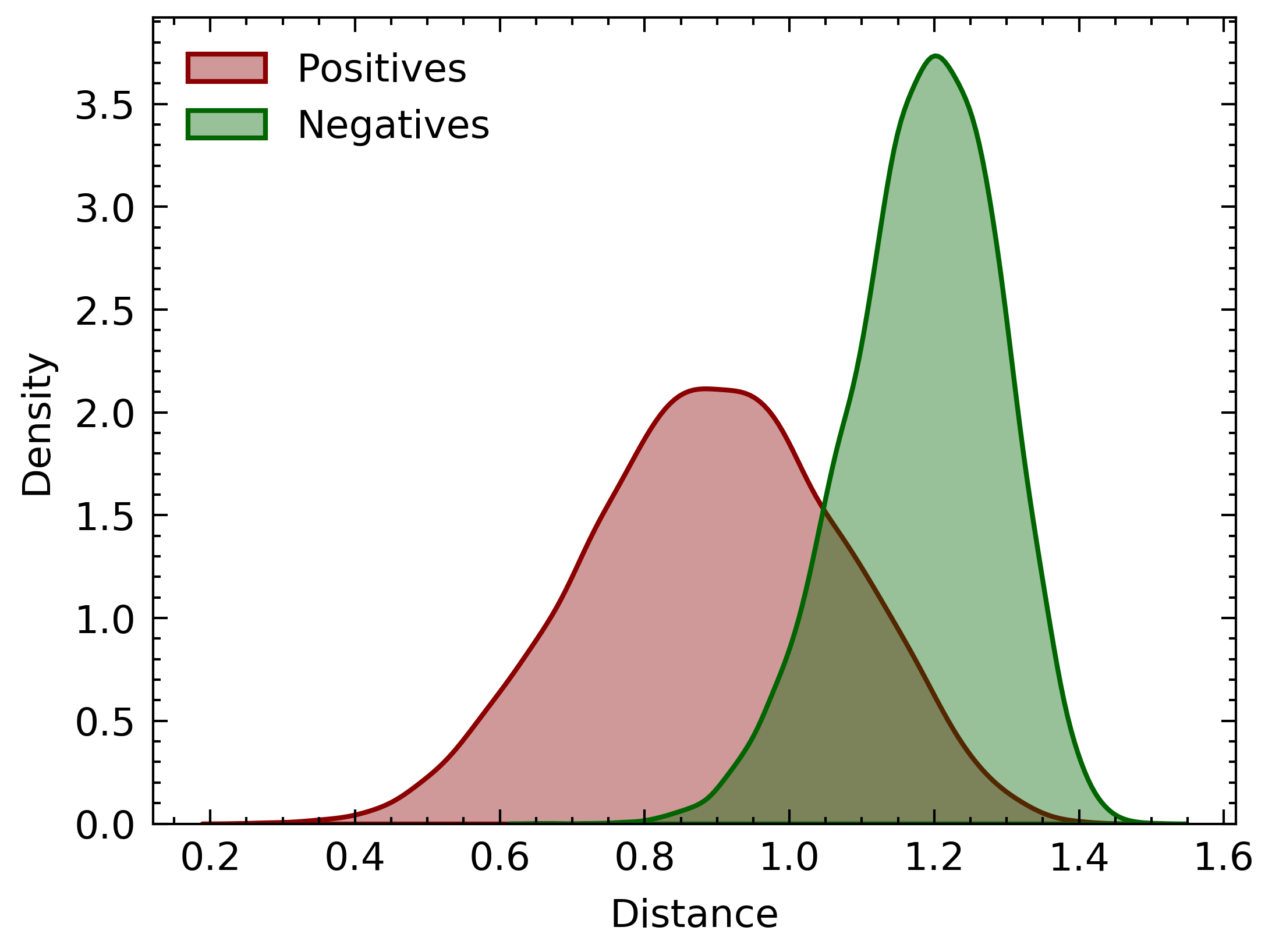} &
		\includegraphics[width=0.23\linewidth]{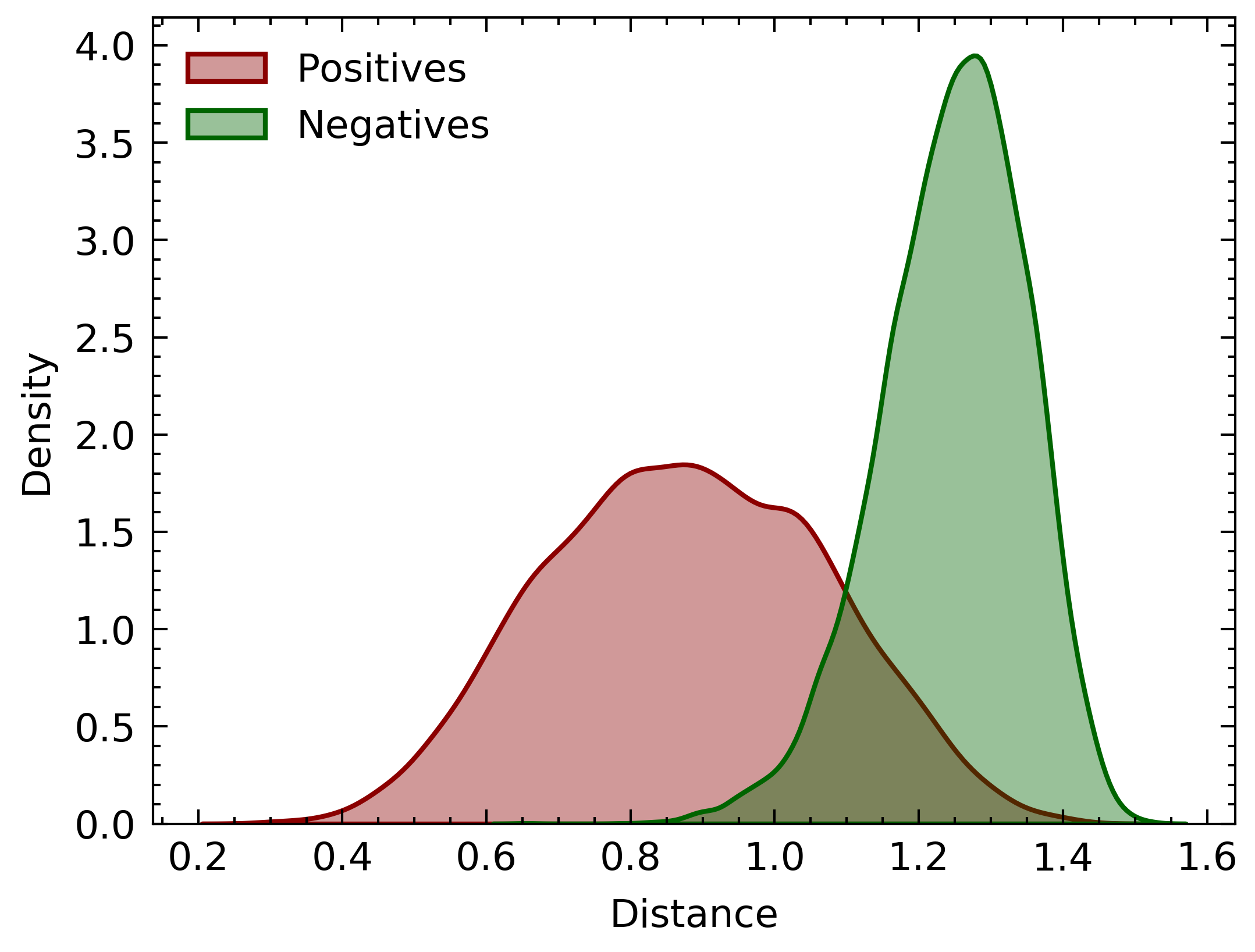} &
		\includegraphics[width=0.23\linewidth]{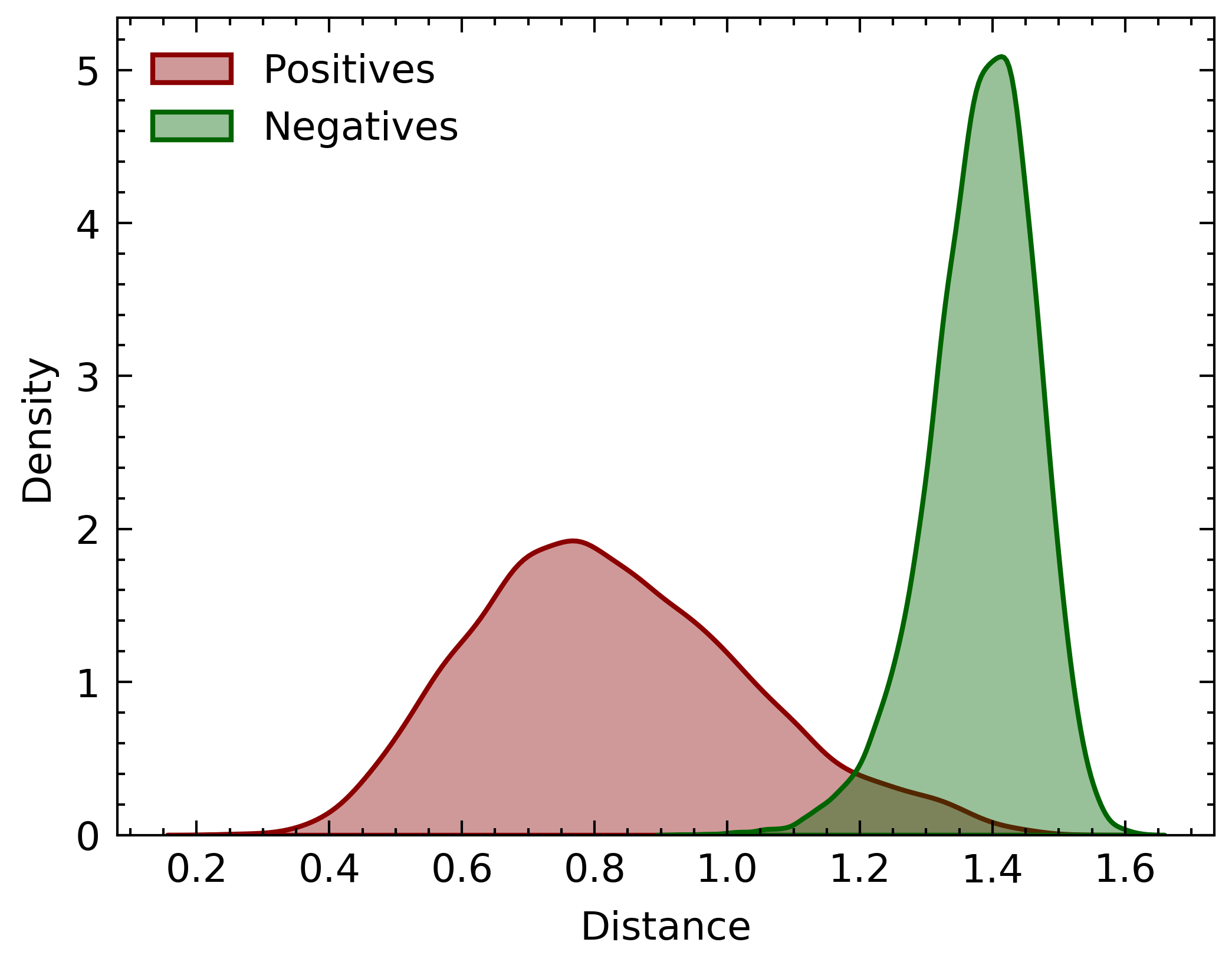} &
		\includegraphics[width=0.23\linewidth]{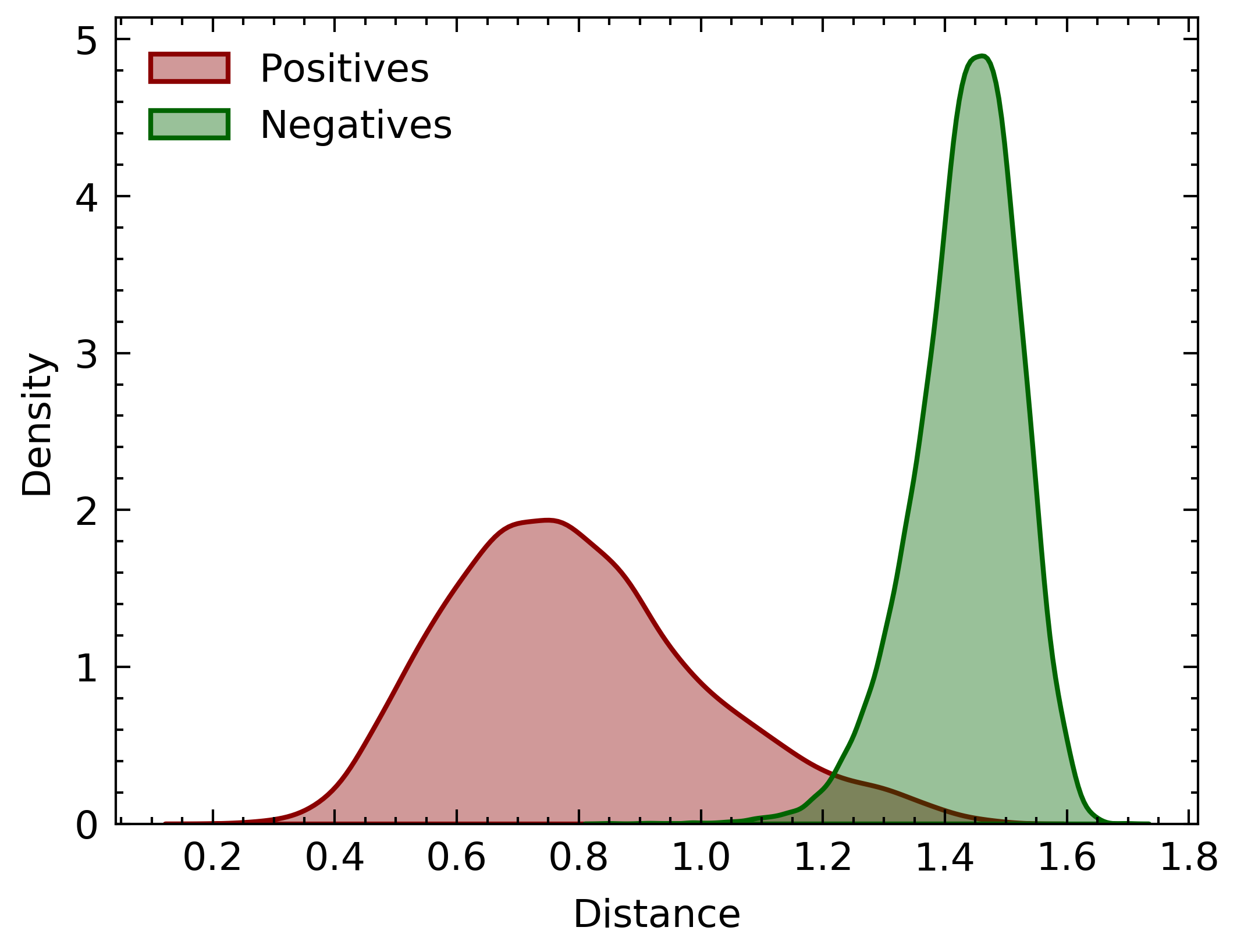} \\
		~\\
		(a) Direct transfer & (b) Baseline & (c) IDM & (d) IDM++
	\end{tabular}
	\caption{Target domain's distribution of different methods when transferring from Market-1501 to DukeMTMC-reID. }
	\label{fig:kde_pos_neg}
\end{figure*}

\textbf{Scalability for different backbones.} Our proposed IDM++ can be easily extended to other backbones which are equipped with different normalization layers \cite{pan2018two,li2016revisiting,zhuang2020rethinking}. We take the commonly used IBN \cite{pan2018two} as an example and use the ResNet50-IBN \cite{pan2018two} as a stronger backbone. As shown in Fig.~\ref{fig:ibn_map_r1}, our method can achieve the consistent performance gains with the stronger backbone and the performance gain is especially obvious on the largest and most challenging MSMT dataset.

\subsection{Analysis and Discussions}
\label{sec:analysis}
\textbf{Analysis on pseudo labels quality.}
Our method can also be included into the pseudo-label-based UDA re-ID methods \cite{dai2020dual,ge2020self} where the quality of pseudo labels is important during training, because target training data is supervised by pseudo labels. Following the exisiting methods \cite{dai2020dual,yang2020asymmetric,ge2020self}, we use both the BCubed F-score \cite{amigo2009comparison} and the Normalized Mutual Information (NMI) score to evaluate the quality of pseudo labels. For F-score and NMI score, the score towards 1.0 implies better clustering quality and less label noise. As shown in Fig. \ref{fig:pseudo_labels}, we visualize the F-score, NMI score, and Rank-1 score when training on Market-1501 and testing on MSMT17. It shows that our IDM++ can outperform our IDM and Baseline method by a large margin no matter the quality of pseudo labels or on the Rank-1 performance.

\textbf{Analysis on the bridge and alignment of domains' distribution.} In Fig.~\ref{fig:kde_all}, we provide the visualization on the bridge between intermediate domains and the source/target domain, and the distribution alignment between domains when transferring from the source domain Market-1501 to the target domain DukeMTMC-reID. In Fig.~\ref{fig:kde_all} (a), the distance distribution $d(P_{s},P_{\rm inter})$ between the source and intermediate domains is almost aligned with $d(P_{t}, P_{\rm inter})$ between the target and intermediate domains. The little discrepancy between $d(P_{s},P_{\rm inter})$ and $d(P_{t}, P_{\rm inter})$ shows the distribution of learned intermediate domains is diverse enough to bridge source and target domains. If the learned intermediate domains are not diverse or are dominated by either domain, the discrepancy of the distribution between $d(P_{s},P_{\rm inter})$ and $d(P_{t},P_{\rm inter})$ will be large. To evaluate the effectiveness of our proposed cross-domain consistency loss that enforces the consistency between source/target samples and their mirrors generated by the MGM, we randomly sample 20,000 sample pairs in each domain and calculate the Euclidean distances of their L2-normalized features to visualize the distributions in Fig.~\ref{fig:kde_all}~(b) and (c).
As shown in Fig.~\ref{fig:kde_all}~(b), the distribution of source samples and the distribution of their mirrors in intermediate domains can be well aligned. For target samples and their mirrors, their distributions are also well aligned as shown in Fig.~\ref{fig:kde_all}~(c).
It shows that our proposed MGM can well preserve the original source / target identity-discriminative distribution in intermediate domains.
By aligning source and target domains to intermediate domains respectively, distributions of source and target domains can be well aligned as shown in Fig.~\ref{fig:kde_all}~(d).

\textbf{Analysis on the discriminability of the target domain's features.} In Fig.~\ref{fig:kde_pos_neg}, we visualize on the distribution of target domain's positives and negatives and compare among four methods including Direct transfer, Baseline, our IDM, and our IDM++. Specifically, we randomly sample 10,000 positive pairs and 10,000 negative pairs in the test-set of the target domain when transferring from Market-1501 to DukeMTMC-reID. When the overlap between positives' and negatives' distributions is smaller, the model will be more discriminative. As shown in Fig. \ref{fig:kde_pos_neg}, when training with our IDM or IDM++, the discriminability gain is significant than the Direct transfer or the Baseline method.

\section{Conclusion} \label{sec:conclusion}
This paper provides a new perspective for cross-domain re-ID: the source and target domain are not isolated, but are connected through a series of intermediate domains. This perspective motivates us to align the source / target domain against these shared intermediate domains, instead of the popular source-to-target alignment. To this end, we propose an IDM and an MGM module and integrate them into IDM++. 
IDM first discovers the intermediate domains by mixing the source-domain and target-domain features within the deep network and then minimizes the  source-to-intermediate and target-to-intermediate discrepancy. 
MGM further reinforces the domain alignment by solving a side effect of IDM. Specifically, the intermediate domains generated by IDM lack the original identities in the source / target domain, while MGM compensates for this lack by mapping the original identities into the IDM-generated intermediate domains.
Both IDM and MGM can be seamlessly plugged into the backbone network and facilitate mutual benefits. Experimental results show that IDM significantly improves UDA baseline and MGM brings another round of substantial improvement. Integrating IDM and MGM, IDM++ achieves the new state of the art on both UDA and DG scenarios for cross-domain re-ID.

\ifCLASSOPTIONcaptionsoff
  \newpage
\fi

\bibliographystyle{IEEEtran}
\bibliography{IEEEabrv,ref}
\end{document}